\documentclass[11pt,a4paper, table]{article}
\usepackage[hyperref]{eacl2021}
\usepackage{times}
\usepackage{latexsym}

\usepackage{graphicx}
\usepackage{hhline}
\usepackage{highlight}
\usepackage{highlight2}
\usepackage{multirow, makecell}
\usepackage{float}
\usepackage{amsfonts,amsmath,amssymb}
\usepackage{todonotes}
\usepackage{lipsum,graphicx,subcaption}
\usepackage{microtype}

\aclfinalcopy %

\title{An Evaluation Protocol for Generative Conversational Systems}

\author{Seolhwa Lee\thanks{Work performed while at Johns Hopkins University.}  \\
  Korea University \\
  Seoul, South Korea \\
  \texttt{whiteldark@korea.ac.kr} 
 \\\And
 Heuiseok Lim\\
  Korea University \\
  Seoul, South Korea \\
  \texttt{limhseok@korea.ac.kr} \AND
  Jo\~ao Sedoc $^*$\\ 
New York University\\
  New York, USA \\
\texttt{jsedoc@stern.nyu.edu}}

\date{}

\begin{document}
\maketitle
\begin{abstract}
There are a multitude of novel generative models for open domain conversational systems; however, there is no systematic evaluation of different systems.
Systematic comparisons require consistency in experimental design, evaluation sets, conversational systems and their outputs, and statistical analysis.
In this paper layout a protocol for the evaluation of conversational models using head-to-head pairwise comparison. We analyze ten recent models which claim state-of-the-art performance using a paired head-to-head performance (win-loss-tie) on five evaluation datasets.
Our findings show that DialoGPT and Blender are superior systems using Bradley-Terry model and TrueSkill ranking methods. These findings demonstrate the feasibility of our protocol to evaluate conversational agents and evaluation sets. 
Finally, we make all code and evaluations publicly available for researchers to compare their model to other state-of-the-art dialog models.
\end{abstract}

\section{Introduction}
There has been a flurry of recent work in open domain conversational systems which can ideally converse about any topic~\citep{csaky2019deep,gao2019neural}. 
Recent generative conversational systems use end-to-end trained neural network encoder-decoder models~\citep{vinyals2015neural}. Evaluating the improvement between models is difficult as different system are rarely compared to other state-of-the-art models using the same evaluation datasets with the same evaluation setup. Arguably, the lack of standardized comparisons of systems impedes progress in the field.

The evaluation of generative conversational systems is challenging due to a lack of automatic metrics ~\cite{li2017adversarial,lowe2017towards}. 
For this reason, human evaluation is standard practice.  
Although there is a tremendous amount of research in generative conversational systems recently, there are no standard experimental design or evaluation methods.  This is a crucial issue. A systematic evaluation requires the exact same 1) experimental design, 2) evaluation datasets 3) models and their response utterances, and 4) statistical analyses. To solve this issue, we present an evaluation protocol {\it with code} and template evaluation.

We call for an evaluation protocol and provide a partial solution. In order to forward this, we present a full work through of our proposal by examining an experimental design on various models and evaluation sets, and then we present a thorough analysis of results.  
Specifically, we use a next utterance generation task through the ChatEval\footnote{\url{http://chateval.org}\label{footn:chateval}} A/B paired comparison approach~\cite{sedoc-etal-2019-chateval}. ChatEval has a standard experimental design and evaluation datasets. 

One limitation with this experimental design is the lack of interactive evaluation. There is a trade-off between truly interactive evaluation (or deployment A/B testing) and statistical significance. As seen in \citet{Venkatesh2018}, thousands of interactive conversations are required for statistical significance. While head-to-head next utterance generation comparison is further from the end-task of conversation, it has more statistical power due to the fact that it supports paired tests.  
\citet{Novikova2018} found that relative rankings yield more discriminative results than absolute assessments when evaluating natural language generation. 

In this setting, dialog systems can be viewed as addressing a natural language generation task and {\it not} as being an interactive agent that carries the conversation further. Arguably, due to their lack of planning and reasoning abilities, many current dialog systems are actually sentence level language models. Although next utterance generation is a more artificial task, \citet{logacheva2018dataset} observed a Pearson correlation of 0.6 between conversation-level and utterance-level ratings.

In order to analyze our evaluation protocol,
we perform a large-scale human evaluation of {\bf ten} baseline and state-of-the-art systems. 
We find that while this is a $O(kn^2)$ problem ($k$ evaluation sets and $n$ systems), it is not prohibitively expensive.\footnote{The total cost of all crowdsourcing experiments was approximately \$1,300.} 
For our evaluation protocol: 1) We evaluate ten state-of-the-art models under the same evaluation conditions, 2) we use publicly available evaluation datasets with both single-turn and multi-turn conversational prompts 
and 3) we carefully analyze human annotations and multiple single-turn and multi-turn evaluation sets. 

\section{Evaluation Methodology}
We utilize the ChatEval A/B paired testing framework to evaluate all the systems in our study. Subsequently, we analyze the results using win scores and system ranking analyses.

\subsection{ChatEval A/B Paired Test} \label{subsec:chateval}
ChatEval~\cite{sedoc-etal-2019-chateval} is an online platform that compares responses of two different systems under the same dialog context. ChatEval interfaces with Amazon Mechanical Turk\footnote{\url{https://www.mturk.com/}} to assess models (See the Appendix~\ref{sec:app_chateval} for further details).   
Annotators are presented the next utterance in a conversation given the context of some number of previous turns.
Then, the annotator (i.e., crowd worker) decides which answer (i.e., two possible responses A/B.) is the best or if there is a tie between both systems. The platform randomly distributes the tasks among different annotators allowing an unbiased pairwise evaluation. 
Table \ref{tab:chateval} illustrates the different models, responses as A and B, given the prompt.

\begin{table}[h]
    \centering
    \scalebox{0.9}{
    \begin{tabular}{l}
         \textbf{Prompt:} \textit{Is the sky blue or black?}\\ \hline
            \textbf{A:} \textit{It's a black sky.}\\
            \textbf{B:} \textit{The sky is blue because of an optical effect } \\
                        \textit{known as Rayleigh scattering.} \\ \hline
    \end{tabular}}
    \caption{The example of Chateval A/B paired test.}
    \label{tab:chateval}
\end{table}

\subsection{Ranking}
We follow the Workshop on Machine Translation (WMT) ranking score method \citep[Thesis chapter 7.1]{bojar2018english}: 
\begin{align*}
\small
&Major_{score(win)} = \frac{win}{win+loss} ,  \\
&Major_{score(loss)} = \frac{loss}{win+loss} , \\
&Distinct_{score(win)} = \frac{win}{win+loss+tie} , \\  
&Distinct_{score(loss)} = \frac{loss}{win+loss+tie} .
\end{align*}
The major score (i.e., \(Major_{score}\)) is for the building pairwise ranking, which accepts only both the {\it win} and {\it loss} count ignoring {\it tie}. On the other hand, the distinct score (i.e., \(Distinct_{score}\)) includes {\it tie} to assess how frequently the systems were judged to be better than or equal to the others. This penalization allows one to differentiate systems more carefully. We also consider the total system win count (i.e., frequency) as rank method, for example if Blender ``wins'' over DialoGPT and ConvAI2, then its system win count is two. 

Furthermore, we analyze our system comparisons using two standard statistical ranking methods: TrueSkill~\cite{herbrich2007trueskill} and Bradley-Terry (BT) model~\cite{bradley1952rank}. TrueSkill is a non-parametric online algorithm to evaluate a relative skills of players through the competitions such as Microsoft's Xbox Live. 
For TrueSkill, we follow the WMT-TrueSkill~\cite{sakaguchi2014efficient} approach, which is used for ranking MT systems in WMT by measuring the `relative ability' from the space of system pair matchings.
The Bradley-Terry model is a parametric probability model that can predict the outcome of a paired comparison. We carry out experiments with these different methods.

\section{Evaluation Datasets} \label{sec:data}
We evaluate generative conversational systems using four datasets as single-turn (i.e., NCME, DBDC, Twitter, Cornell Movie DC) and one dataset (ESL) for the multi-turn evaluation. This allows us to compare between single-turn and multi-turn capability of each model. For the evaluation on the multi-turn datasets we only use models that can capture conversational history (DialoGPT, Blender, CakeChat (HRED implementation), ConvAI2 (seq2seq), ConvAI2 (KV-MemNN), ParlAI (controllable)). These evaluation datasets are publicly available on the ChatEval web portal. Aside from Cornell Movie DC evaluation dataset which has 1000 prompts, all other evaluation datasets have 200 prompts.

\paragraph{i. Neural Conversational Model Evaluation set (NCME)} \citet{vinyals2015neural} conducted human evaluation using a hand-crafted set of 200 single turn prompts. A large portion of these prompts are questions. The NCME dataset includes both specific domains, noisy and general domain prompts, such as questions about {\it{morality}} and  {\it{general knowledge in math}}. 

\paragraph{ii. Dialog Breakdown Detection Challenge Evaluation set (DBDC)} The DBDC dataset consists of a series of text-based conversations between a human and a chatbot where the human was aware they were chatting with a computer \cite{higashinaka2016dialogue}. The evaluation set is a selection of 200 single turns from the DBDC 3 dataset~\cite{higashinaka2017overview}.

\paragraph{iii. Twitter} A set of 200 prompts from conversational threads were randomly drawn from the ParlAI \cite{miller2017parlai} Twitter derived test set.

\paragraph{iv. Cornell Movie Dialogue Corpus} The Cornell Movie Dialogue Corpus (DC) \cite{Danescu-Niculescu-Mizil+Lee:11a} contains accurate speaker annotations for each participant's utterances in each conversation. We use 1000 prompts selected by \cite{baheti2018generating}, which extracts two turn conversation as source target pair from original data. 

\paragraph{v. English as a Second Language (multi-turn)} we use the scraped 1000 10-turn conversations between human-human for English as a Second Language learners (ESL) as a multi-turn evaluation.\footnote{\url{http://ESLfast.com}} We selected 200 3-turns snippets from conversations. To the best of our knowledge, our work is the first to use this dataset; however, it is publicly available on ChatEval.

\section{Systems} \label{sec:systems}
We chose systems based on accessibility and reproducibility. 
We evaluate two state-of-the-art models, Blender and DialoGPT. Both are publicly available, unlike other chatbot models such as Meena~\cite{adiwardana2020towards} and GPT-3~\cite{brown2020language}. In addition, we evaluate several other systems:  Controllable dialogue, ConvAI2 (seq2seq, KV-MemNN), Transformer, OpenNMT(Twitter, OS), Cakechat and DC-NeuralConversation. Next, we summarize these systems (see Appendix~\ref{appen:system} for further details):
~\\
\textbf{Human baselines} We use human baselines: NCME human 1, NCME human 2, DBDC human, Twitter baseline and Cornell movie DC baseline.\textsuperscript{\ref{footn:chateval}} NCME  and DBDC have two human baselines which are created post-prompt selection. Whereas Twitter, DBDC, and ESL baselines are from the next turn in the conversation. 
\textbf{ParlAI (Blender)} \cite{roller2020recipes} is recently presented as open-domain generative conversational model from the ParlAI platform \footnote{\url{https://parl.ai/}}. Blender uses a ensemble of various models to create a conversational system. This leads to high quality response generation. Blender achieves the state-of-the-art on existing approaches in multi-turn dialogue yielding humanness and engagingness measurements. Notably, to our knowledge Blender was not compared to DialoGPT until this work.
~\\
\textbf{DialoGPT}
\citep{zhang2019dialogpt} is another state-of-the-art model which uses a GPT framework trained on Reddit data. Its responses have higher performance to the context-consistent response on single-turn dialogue.
~\\
\textbf{ParlAI (Controllable dialogue)} This model is oriented towards controllable generation and has repetition-controlled, inquisitive and interesting responses which obtained the highest human Likert scores in a published study \cite{see2019makes}.
~\\
\textbf{ConvAI2 (seq2seq)} We select this model as a basic baseline of the deep learning approach. ConvAI2 model\footnote{\url{https://github.com/facebookresearch/ParlAI/tree/master/projects/convai2}} from ParlAI is based on the seq2seq model to the ConvAI2 competition\footnote{\url{http://convai.io/}}.
~\\
\textbf{ConvAI2 (KV-MemNN)} ConvAI2 (KV-MemNN) is Key-Value Profile Memory Network~\cite{dinan2019second, zhang2018personalizing} from ParlAI \footnote{\url{https://github.com/facebookresearch/ParlAI/blob/master/projects/convai2/baselines/kvmemnn/interactive.py}} and this model was a baseline for the ConvAI2 competition. We only used this model for multi-turn evaluation.
~\\
\textbf{Transformer} \cite{vaswani2017attention} is most commonly used architectures in generative conversational models these days. We employ conversation data trained Transformer \cite{csaky-etal-2019-improving} as a basic baseline for reflecting generative conversational model.
~\\
\textbf{OpenNMT (Twitter)} is OpenNMT \cite{opennmt} trained model with seq2seq with Attention trained on Twitter dataset from ParlAI.
~\\
\textbf{OpenNMT (OS)} is OpenNMT trained model with seq2seq with Attention trained on OpenSubtitle (OS) questions only.
~\\
\textbf{CakeChat} is a emotional generative dialogue system using Hierarchical Recurrent Encoder-Decoder (HRED) by Replika.ai \footnote{\url{https://replika.ai/}}.
~\\
\textbf{DC-NeuralConversation} \cite{baheti2018generating} is OpenNMT based neural conversation model which implements topic and semantic distributional constraints to improve quality of generated responses.

\section{Results and Analysis}

We first begin by analyzing crowd worker's annotations and evaluation sets then we evaluate systems. The purpose of this is to systematically detail the issues in this evaluation methodology. Finally, we show the coarse-grained approach to measure chatbot quality using paired test results in Section~\ref{subsec:ranking}.

\subsection{Crowd Worker Analysis}
\label{subsec:crowd}
We had three (occasionally more) Amazon Mechanical Turk (AMT) workers to judge each A/B paired test. In general, for every 200 prompts\footnote{Recall that Cornell Movie DC is the exception having 1000 prompts.} and responses there are 600 ratings when 3 voters are employed. The annotation instruction in ChatEval are not specific which \citet{sedoc-etal-2019-chateval} note leads to low inter-annotator agreement (IAA). Throughout our experiments we find Fleiss' Kappa~\cite{fleiss1971measuring} to be between 0.1 and 0.5 which is seemingly unacceptably low;  however, we are able to rank systems with statistical significance. {\it Why?} As \citet{amidei-etal-2018-rethinking,amidei2019agreement} note IAA is not necessary for significance testing. Next, we explored multiple agreement analyses to further understand annotator judgements.
~\\
\textbf{Weak Agreement} We studied the workers voting results for the our experiments using weak agreement, proposed by~\citet{devault2011toward}. This metric for measuring human judge agreement in about 50\% of the cases on the same response for a given prompt. This statistic regards a model response as appropriate when at least one worker prefers the response to the other alternative model's response. We scored this weak agreement as $all_{agree}$, $A/B_{dis}$, $one_{dis}$ and $all_{dis}$ per prompts. Each of these statistics are compared to the agreement of all of workers, in the case of $A/B_{dis}$ ties are excluded (so at least one annotator must prefer the response of model A and another annotator prefer model B). $one_{dis}$ is similar to $A/B_{dis}$ but includes ties. $all_{dis}$ indicates there is at lease one vote for A, another for B as well as a tie. We tabulated these statistics for every model comparison by evaluation dataset in Tables~\ref{tab:ncm2},~\ref{tab:dbdc2},~\ref{tab:twitter2} and~\ref{tab:cornell2} (see those Tables in the Appendix~\ref{appen:weak}).

\begin{figure}[tbh]
    \centering
    \includegraphics[width=0.4\textwidth]{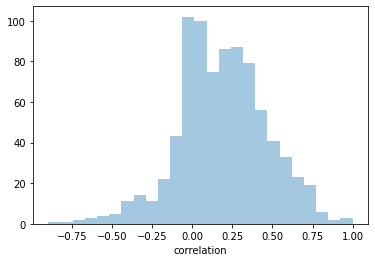}
    \caption{A histogram of the Spearman correlation between one annotator's ratings and the others.}
    \label{fig:annotator_corr}
\end{figure}

In Table~\ref{tab:ncm2}, we analyzed the correlation between $major_{score}$ and agreement statistics to observe the relation. We find  $major_{score(win)}$ correlated positively with $all_{agree}$ and negatively with all of the disagreement scores (i.e., $A/B_{dis}$, $one_{dis}$, $all_{dis}$). As expected, all of the disagreement scores correlated weakly with each other, but these correlations are occasionally not statistically significant (see Figure~\ref{fig:aggre_corr_spear} in the Appendix~\ref{appen:correlation}).
~\\
\textbf{Bad Annotators} Next, we qualitatively reviewed crowd workers. Specifically, we focus on a peculiar result of the comparison between Blender and ConvAI2\footnote{Recall that ConvAI2 is same ParlAI (ConvAI2).} in the NCME (see NCME heatmap in Figure~\ref{fig:heatmap_results}). We find 54 examples where annotators chose the response that was clearly worse.\footnote{The authors verified this manually.}  Hence, even though ConvAI2 is preferred over Blender by 2 \% (see Table~\ref{tab:ncm2} in Appendix~\ref{appen:weak}) this result does not hold. Seven annotators accounted for the 54 examples.  We found that the average Cohen's Kappa as well as correlation to other annotators are negative. Given that there are roughly one thousand annotator in our study, this may be due to chance; however, is does seem to indicate random guessing.
~\\
\textbf{Annotator Correlation}
\citet{amidei2019agreement} argue for a correlation analysis. We studied the overall correlation between the judgement of an annotator on a prompts to all other annotators. Figure \ref{fig:annotator_corr} shows that similar to our qualitative study there are a significant portion of negatively correlated annotators. 

\subsection{Evaluation Dataset Analysis}
One other dimension of interest is prompt validity. Concretely, we wanted to understand if a prompt is useful in assessing the relative quality of chatbot responses. This is quantified using item-total correlations~\citep{henrysson1963correction}. As seen in Figure \ref{fig:prompt_validity}, we found many prompts which are randomly selected from Twitter and Cornell Movie D.C. have low question validity. The DBDC evaluation set has more low validity prompts compared to both NCME and ESL. 

\subsection{Ranking Results} \label{subsec:ranking}
\begin{figure*}
\centering
    \includegraphics[width=0.5\linewidth]{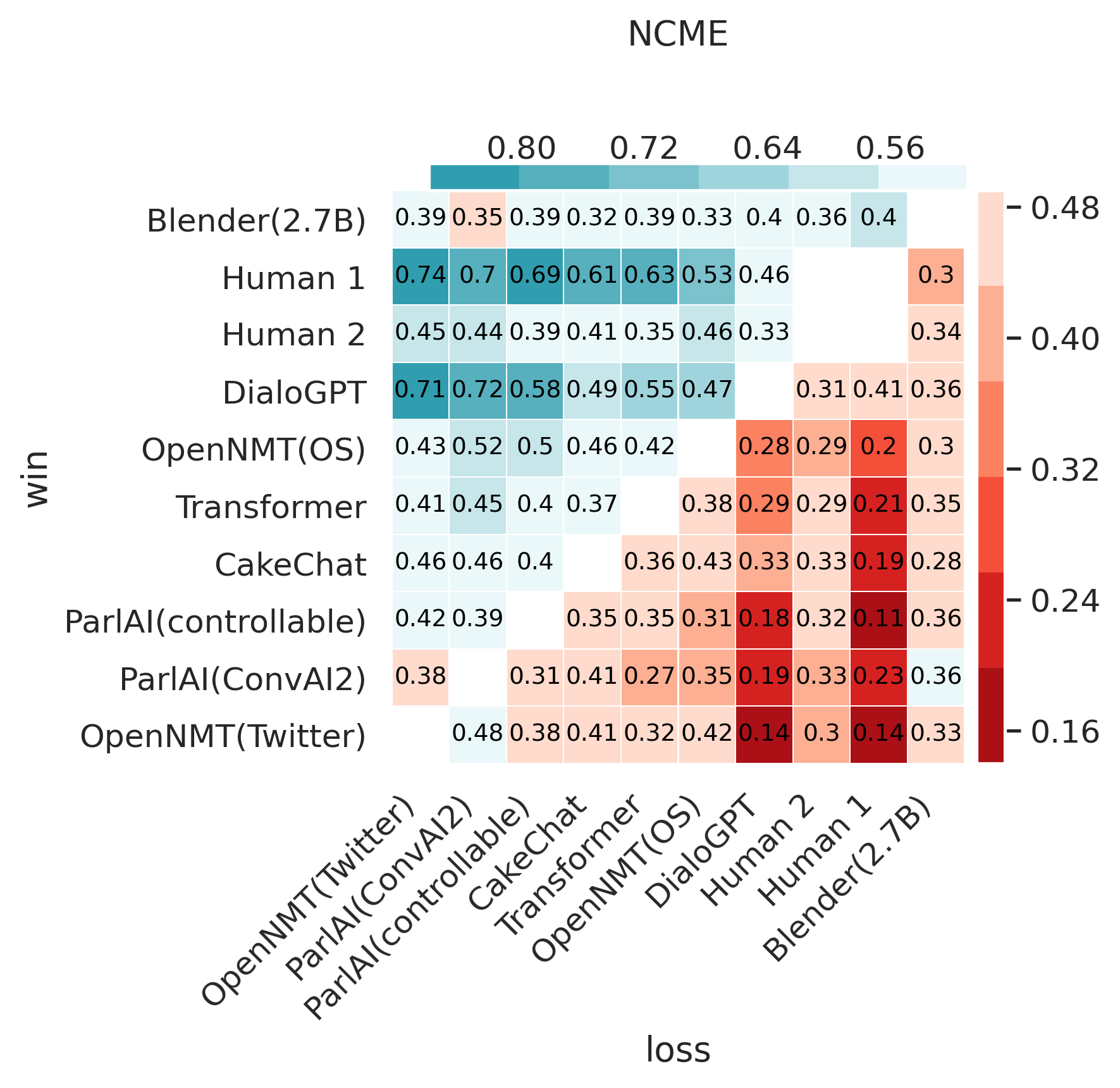}\hfil
    \includegraphics[width=0.5\linewidth]{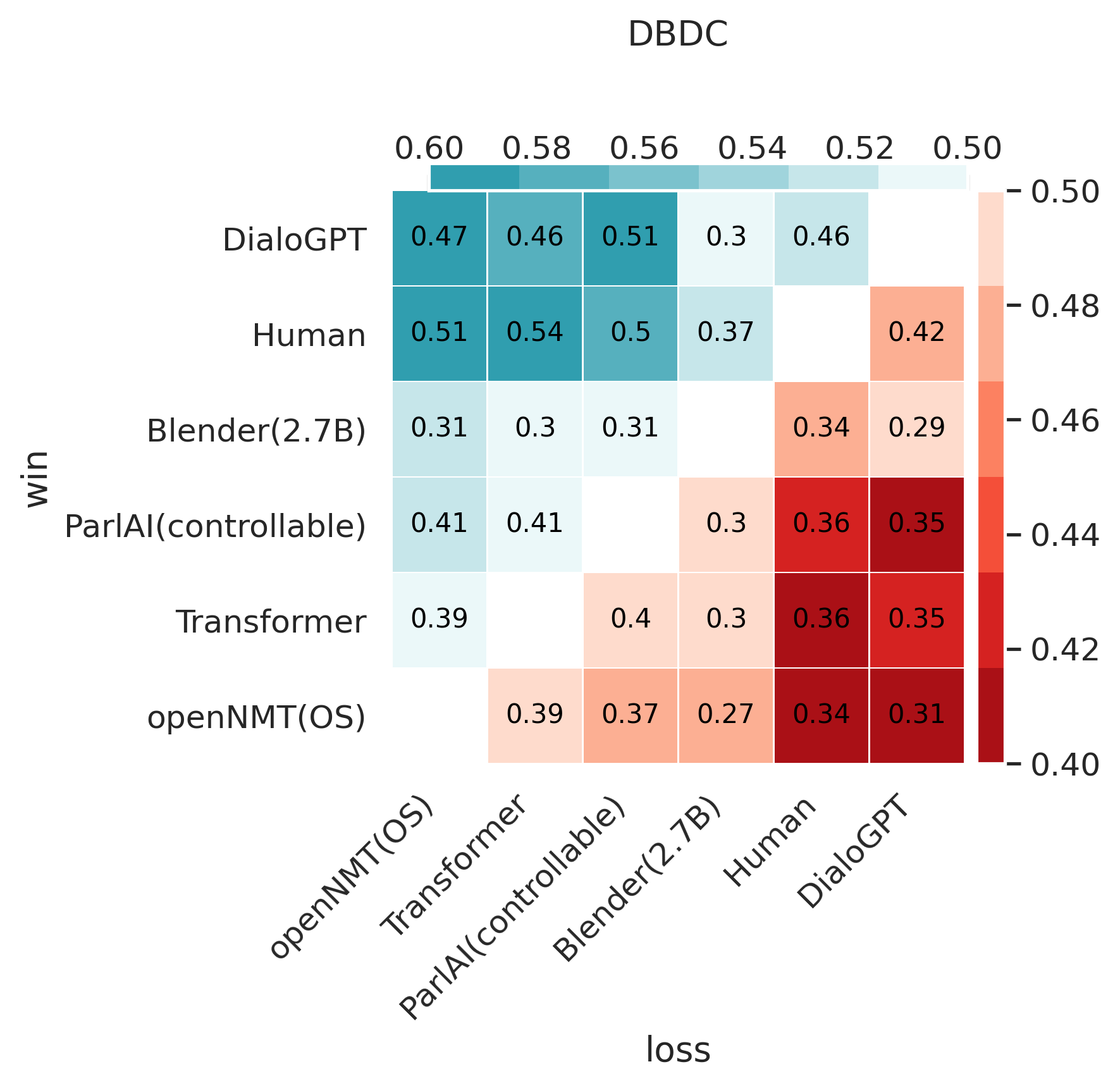}\par\medskip
    \includegraphics[width=0.5\linewidth]{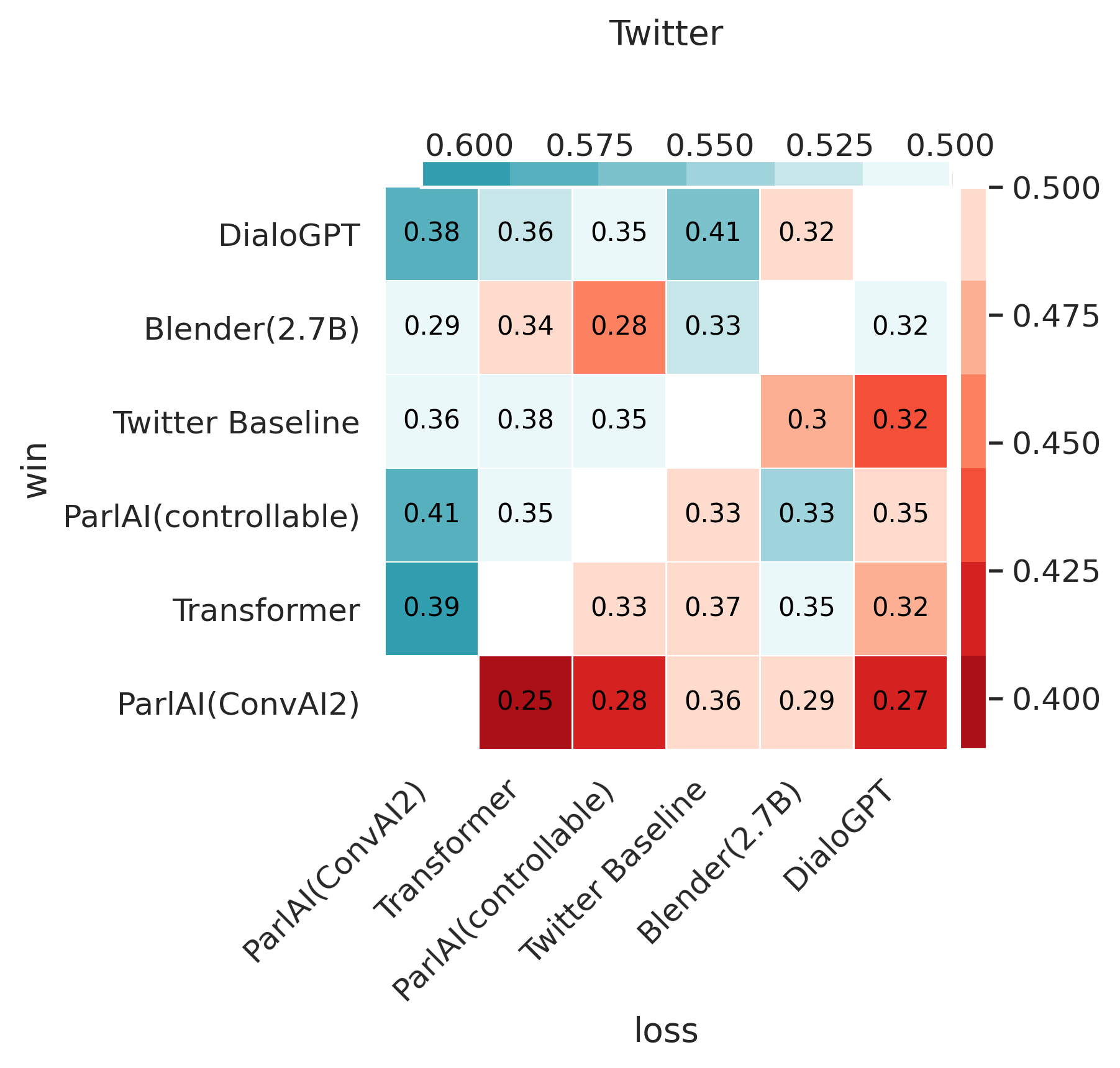}\hfil
    \includegraphics[width=0.5\linewidth]{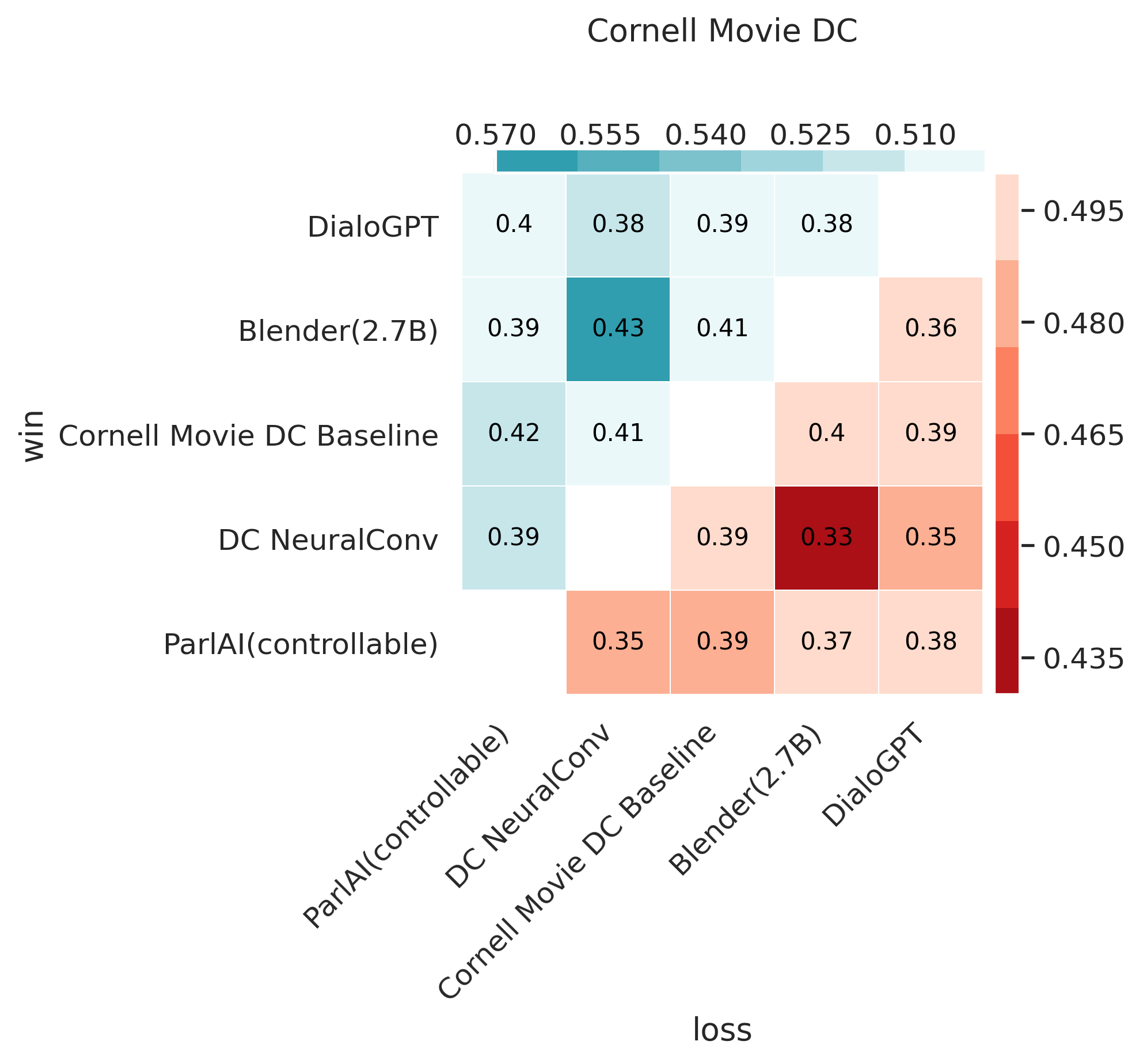}\hfil
    \includegraphics[width=0.5\linewidth]{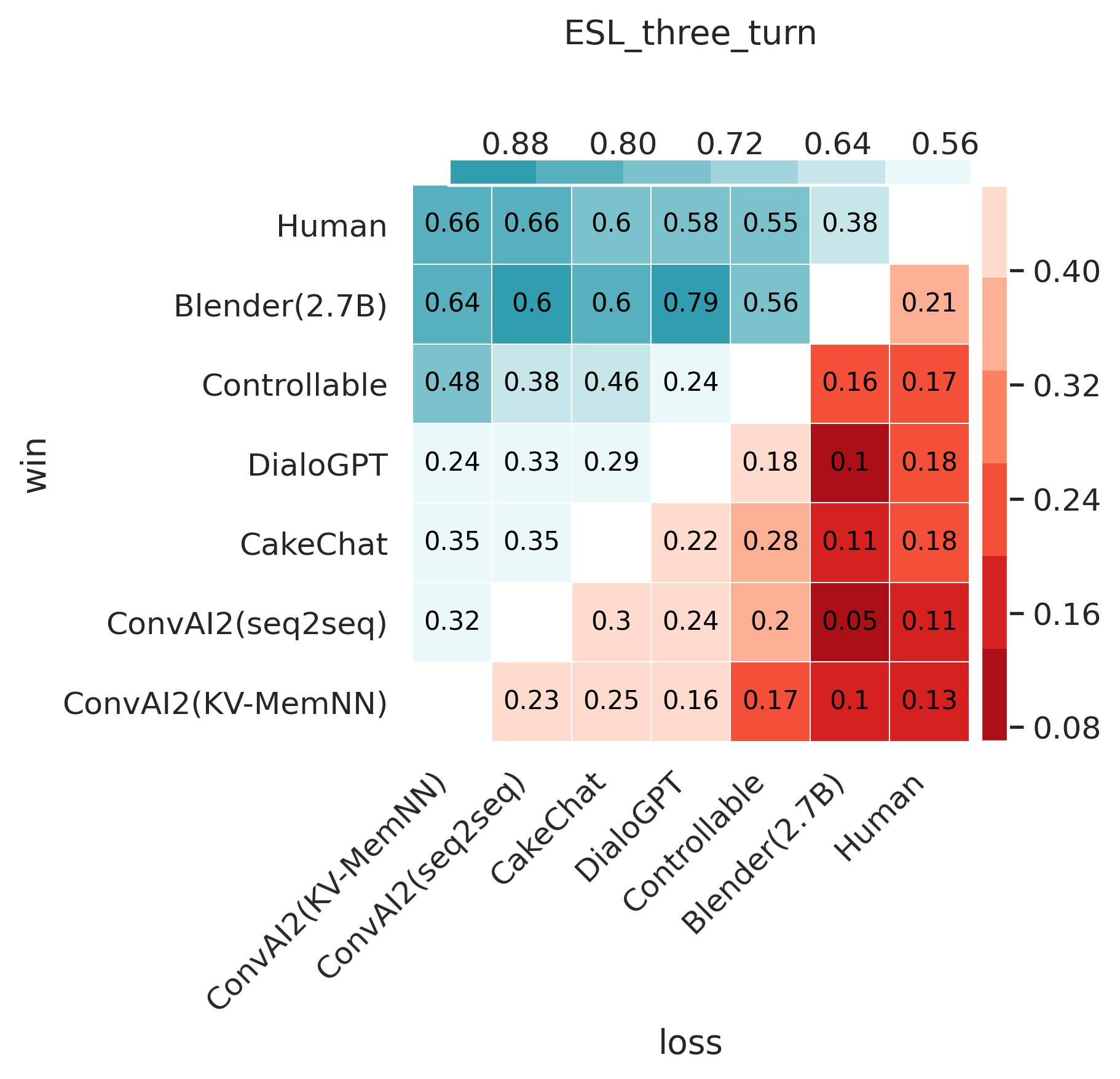}\hfil
\caption{Heatmap for the ratio of A/B model comparison using generative conversational model with single-turn evaluation sets: NCME, DBDC, Twitter, Cornell Movie DC and multi-turn evaluation set: ESL three turns. The gradation of color (blue: wins, red: losses) indicates major score. The cell values are the distinct score. The y-axis indicates wins and x-axis indicates losses. The models are ordered by win count.}
\label{fig:heatmap_results}
\end{figure*}

\begin{figure}[H]
\centering
    \includegraphics[width=1\linewidth]{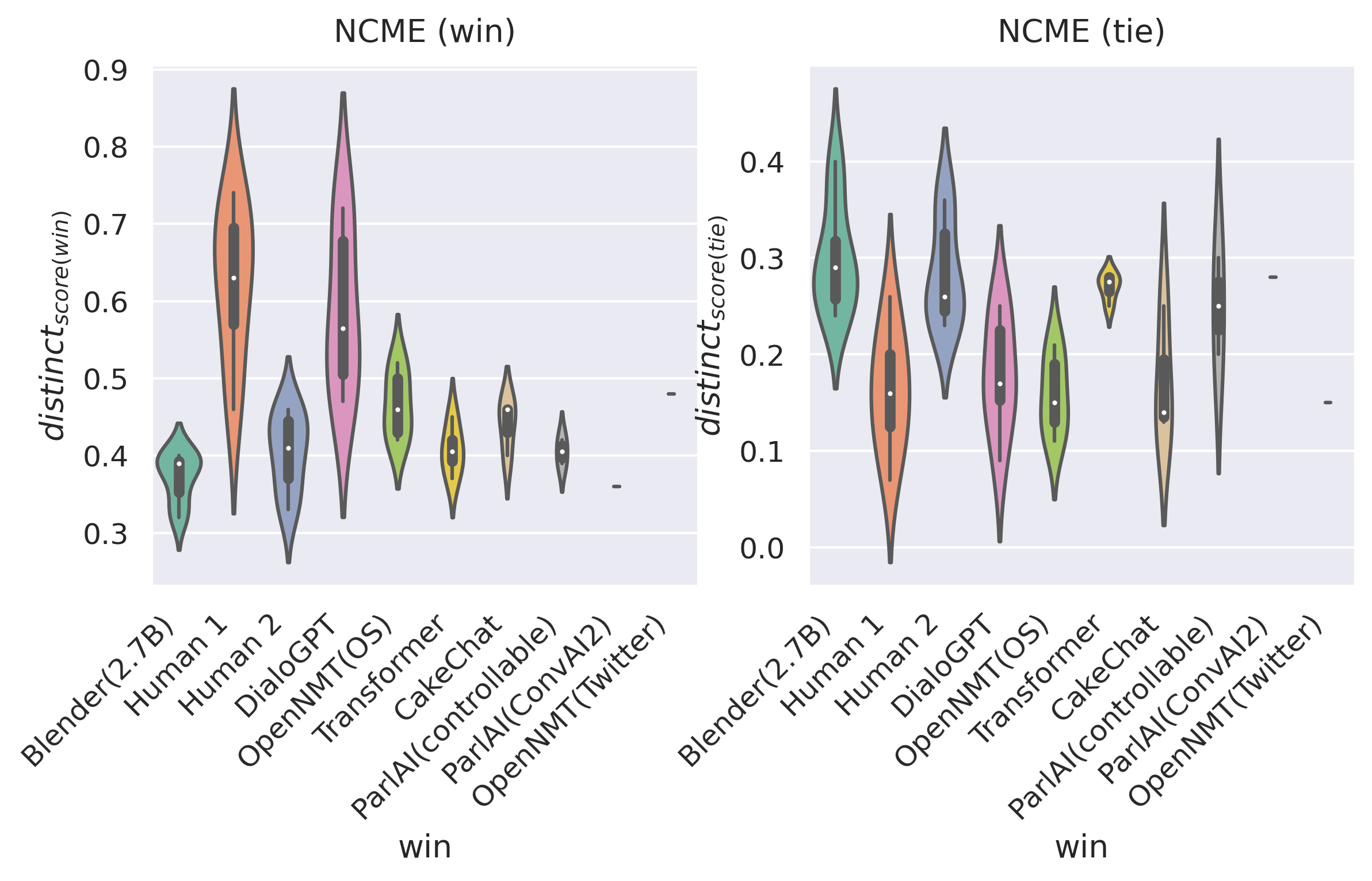}\hfil
\caption{The A/B model comparison results on NCME. Note that the x-axis indicates wins and y-axis indicates $distinct_{score(win)}$ (left) and $distinct_{score(tie)}$ (right).}
\label{fig:ncm_violin}
\end{figure}

Figure~\ref{fig:heatmap_results} shows the evaluation results of model comparisons on the evaluation datasets. We focused on the NCME and ESL evaluations because of the higher question validity; however, all results are available in the Appendix. Furthermore, we explored the winning quality with {\it{tie}} in Figure~\ref{fig:ncm_violin}. We analyzed those results by {\it{frequency}}, {\it{distinctness}} and {\it{majority}}. For the ranking results, we addressed this using {\it{win frequency}}, TrueSkill, and the BT model. 

\begin{figure}[H]
\centering
    \includegraphics[width=0.8\linewidth]{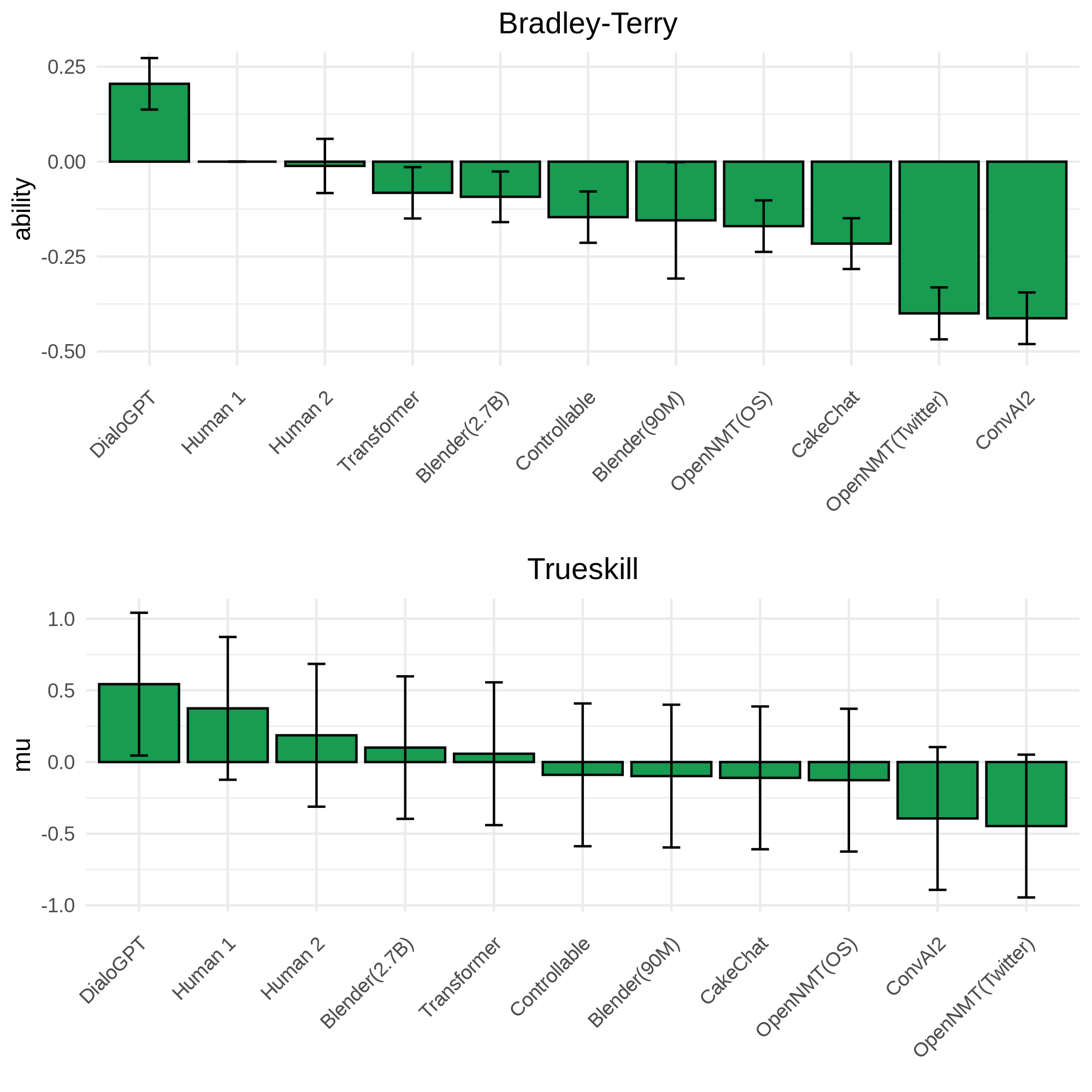}\hfil
\caption{TrueSkill and Bradley-Terry result on NCME. The y-axis indicates ability score (upper) or mu score (bottom) and displays a confidence of the guessed score. The mu score also indicates an average skill of player. The x-axis ordered via score in a descending manner.}
\label{fig:bradley-trueskill}
\end{figure}

~\\
\textbf{Frequency} Blender and DialoGPT had the highest frequency of winning (win count) over all evaluation datasets. NCME in Figure~\ref{fig:heatmap_results} shows the ranking is Blender $>$ NCME human 1 $>$ NCME human 2 $>$ DialoGPT $>$ OpenNMT (OS) $>$ Transformer $>$ CakeChat $>$ ParlAI (controllable) $>$ ConvAI2 (seq2seq)\footnote{Note that ConvAI2 (seq2seq) is same ParlAI (ConvAI2).} $>$ OpenNMT(Twitter). 
Blender was preferred over every models except ConvAI2 in evaluation datasets.\footnote{In Section~\ref{subsec:error} we performed a qualitative investigation of Blender vs ConvAI2.} Nonetheless, DialoGPT ranked highest on multiple single-turn datasets except NCME as seen in Figure~\ref{fig:heatmap_results}.  Generally, both Blender and DialoGPT performed statistically significantly better than the other systems. 
~\\
\textbf{Distinctness} In Figure~\ref{fig:ncm_violin}\footnote{ConvAI2 and OpenNMT(Twitter) each only win once and thus the there is only one observation.}, Blender had \(distinct_{score(win)} < 0.5\), but DialoGPT remarkably showed higher \(distinct_{score(win)} > 0.8\) with NCME Human 1 in NCME (win), although Blender is preferred to DialoGPT. In other words, Blender has a small scale of variance, on the other hand, DialoGPT has a large scale of variance. We find this results are similar on other single-turn evaluation datasets (see Appendix~\ref{appen:violin}).
Blender has a higher \(distinct_{score(tie)}\) than DialoGPT, which can be interpreted as more distinctness rather than Blender. The wider sections of the plot represent a higher probability that members of the population will take on the given value. In contrast, the skinnier sections represent a lower probability. Controllable and CakeChat have a lower probabilities on tie, which means \(distinct_{score(tie)}\) ranges have diversity. But Transformer has nearly static \(distinct_{score(tie)}\) in every comparisons on NCME. 
~\\
\textbf{Majority} NCME showed both human 1 and DialoGPT are highest \(major_{score} \geq 0.8\) in Figure~\ref{fig:heatmap_results}. Specifically, DialoGPT shows strong results in seq2seq models (i.e., ConvAI2, Controllable). On the other datasets in Figure~\ref{fig:heatmap_results}, we found DialoGPT has still better \(major_{score}\) than the other models. \newline
\textbf{TrueSkill \& Bradley-Terry} We further compared the rank from TrueSkill and Bradley-Terry methods. Figure~\ref{fig:bradley-trueskill} shows the same ranking result from 1 to 3 rank and almost similar ranks in each other. Specifically, Blender ranked 4 in TrueSkill but 5 in the BT model. However, we found that the ranks are different between Figure~\ref{fig:heatmap_results}'s NCME and Figure~\ref{fig:bradley-trueskill}, but still DialoGPT and Blender are higher rank, although Transformer ranked higher than Blender in BT. This result is similar on other evaluation datasets (see Appendix~\ref{appen:ranking}). We find the BT model predicts lower standard error than TrueSkill. This is likely due to the parametric nature of the BT model.

\subsection{Error Analysis} \label{subsec:error}

\textbf{Lexical Diversity \& Length} We use the distinct-1 and distinct-2 metrics~\cite{li2015diversity} for measuring the lexical diversity in models responses. The distinct-\(n\) is the number of unique \(n-gram\) in the model's responses divided by the total number of generated tokens. 

\begin{figure}[H]
    \centering
    \scalebox{0.7}{
    \begin{tabular}{|c|c|c|c|}
    \hline
         {} & {Distinct-1} & {Distinct-2} & {Avg.sent.length}\\ \hline
                {Blender (2.7B)} & {0.28} & {0.60} & {16.3} \\ \hline
                {Blender (90M)} & {0.11} & {0.31} & {21.9} \\ \hline
                {DialoGPT} & {0.23} & {0.51} & {8.1} \\ \hline
                {NCME Human 1} & {0.31} & {0.42} & {2.9} \\ \hline
                {NCME Human 2} & {0.36} & {0.68} & {5.4} \\ \hline
    \end{tabular}}
    \caption{The results of distinct-1, distinct-2 and average sentence length of NCME.}
    \label{tab:distinct} 
\end{figure}

We found that response length may correlate with human judgements on NCME but notably ConvAI2(KV-MemNN) model has longer responses but worse overall score. In Figure~\ref{fig:heatmap_results}, we discovered the NCME human 1 and DialoGPT have higher \(major_{score}\) with short average sentence lengths in Table~\ref{tab:distinct}. In contrast, the human 2 and Blender have longer one. But the distinct-1 and distinct-2 between Blender (2.7B) and DialoGPT are not significant difference.
~\\
\textbf{Model Size} We found that indeed model size matters. Blender (90M) loses Blender (2.7B) and also has longer responses (this result can see in Table~\ref{tab:ncm2}), although Blender (2.7B) responses quality is better than Blender (90M). We assumed the response quality related to distinct-\(n\) by observing Blender's results.
~\\
\textbf{Blender with Persona} We investigated the results between Blender with persona-free and Blender with persona for identifying persona affect the result. The persona  with chit-chat have been used for resolving the problem of the lack of a consistent personality~\cite{li2016persona}, which is trained over many dialogs each with different speakers.

Table~\ref{tab:persona} describes persona chat example between Blender with persona and persona-free in NCME (see in Appendx~\ref{appen:persona}). We find Blender with persona model responses using their persona such as example of prompt 1 and prompt 2. As shown in Table~\ref{tab:ncm2}, we find that the persona-free model responses are favored by annotators. This is likely due to heavily persona biased responses. 
Interestingly, we found that $A/B_{dis}$ count higher than most of other comparisons, which may indicate disagreement about the value of the persona versus coherence. 
~\\
\textbf{Qualitative analysis} 
Table~\ref{tab:qualitative-exam} shows the actual responses given the prompt. The responses across each model are noteworthy and reflect their characteristic. Blender responses are much longer than those of other models and are also more coherent. In contrast, DialoGPT and CakeChat did not respond correctly and used a general/safe ``I'm not sure what" pattern, although DialoGPT states the ``morality". 
ConvAI2 also seems to reflect persona without little conversational coherence.                                                              
\begin{table}[h]
    \centering
    \scalebox{0.7}{%
    \begin{tabular}{c|l} \hline
            \multicolumn{2}{l}{\textbf{Prompt} : look, i need help, i need to know more about morality.}\\ \hline
            \textbf{NCME human 2} & I am not a moral person. \\
            \textbf{Blender (2.7B)} & What kind of help do you need? I might be able  \\
             & to help you. I'm a psychologist. \\
            \textbf{DialoGPT} & I'm not sure what you mean by morality. \\
            \textbf{ConvAI2} & I hard work in a factory every day. \\
            \textbf{CakeChat} & I'm not sure what to do. \\
            \textbf{Controllable} & You must be very intelligent. \\
            \textbf{Transformer} & Well I'm thinking of buying a new bedroom set. \\ \hline
    \end{tabular}}
    \caption{\label{tab:qualitative-exam} The actual responses with different dialog systems in NCME.}
\end{table}
~\\

\subsection{Multi-turn Results and Analysis} \label{subsec:multi}

We found the rank Human $>$ Blender $>$ Controllable $>$ DialoGPT $>$ CakeChat $>$ ConvAI(seq2seq) $>$ ConvAI2(KV-MemNN) in Figure~\ref{fig:heatmap_results}'s ESL. Human and Blender have higher score than others. The interesting result is that DialoGPT showed worse performance in multi-turn evaluation. However, Blender continued to be preferred as seen in TrueSkill \& Bradley-Terry (see Appendix~\ref{appen:ranking}). Blender is superior to DialoGPT in the multi-turn evaluation set and the model more efficiently utilized the conversational history.

Distinctness showed quite different than Figure~\ref{fig:ncm_violin} in Figure~\ref{fig:multi_violin}. Blender's $distinct_{score(win)}$ was higher than DialoGPT. 

Figure~\ref{appen:correlation} shows $major_{score}$ correlates with $all_{agree}$ and negatively correlated with all of the disagreement scores except only $A/B_{dis}$ (this case is not statistically significant).

\section{Related Work}

Evaluation of neural dialog generation models models is difficult due to their open-ended nature ,with many possible answers. Therefore, the standard metrics for machine translation or question-answering tasks are not adequate for evaluating such dialogue and also correlate poorly with human judgements \cite{novikova2017we, liu2016not}. 

\citet{li2019acute} propose \textsc{ACUTE-EVAL}, which is the human evaluation technique considering the optimization of the questions for robust measurements over four types of questions: engagingness, interestingness, knowledge and humanness. \textsc{ACUTE-EVAL} has the flow of comparing two full dialogues (i.e., multi-turn dialogues), where a human judge is required to turn their attention to only one speaker within each, and produce a pairwise judgement. Also, \textsc{ACUTE-EVAL} sets up in self-play model chat for the cheaper and faster tests. They provide an explicit benchmark seven ParlAI models of comparison between recent state-of-the-art generative and retrieval models on two tasks, which are Wizard of Wikipedia~\cite{dinan2018wizard} and PersonaChat~\cite{zhang2018personalizing}. However, we conduct the further comparison with both multiple single-turn and multi-turn evaluation through the  ten benchmark models unlike \textsc{ACUTE-EVAL}. Furthermore, we show systematically the diverse of aspects in lexical diversity \& length, personachat and the bunch of ranking method using the score and TrueSkill \& Bradley-Terry. Specifically, we show the deep analysis for understanding annotation quality through the visualization of the results unlike previous work.

Despite of emerging neural dialog generation models, there are still rarely compared to other state-of-the-art models for shared tasks except only ConvAI~\cite{burtsev2018first} and DSTC-7~\cite{d2020overview} challenges. The recent ConvAI challenge is the NeurIPS 2018 ConvAI2 challenge~\cite{dinan2019second}, which is the task for the PersonaChat~\cite{zhang2018personalizing}. PersonaChat is a chitchat dialogue task involved between two participants (human-bot or two humans). Each of them given a persona as a short collection of personal traits. On this challenge, first, the competitor's models were evaluated for automatic metrics, and then conducted human judgement through human-bot chats given the question "How much did you enjoy talking to this user?" on a Likert scale of 1 to 4.

DSTC-7 challenge has three tracks aimed to explore the problem of accurate end-to-end dialog systems and building robust. The dialog generation task is the generation of informational responses grounded in external knowledge (i.e., sentence generation task) in DSTC-7. This task evaluates the competitor's model using both the automatic metrics such as BLEU~\cite{papineni2002bleu} and human evaluation, which evaluates system response in aspect of relevance and interest using crowdsourcing. Human evaluation also were scored on a five Likert scale.

\section{Conclusion}
We lay out an evaluation protocol for generative conversational models and provide a careful analysis of results.
We use multiple models and multiple single-turn and multi-turn evaluation datasets. We also analyze the crowdworkers as well as the evaluation sets. The results show that we can effectively and easily compare systems.

\section*{Acknowledgments}

This research was supported by the MOTIE (Ministry of Trade, Industry, and Energy) in Korea, under the Fostering Global Talents for Innovative Growth Program (P0008749) supervised by the Korea Institute for Advancement of Technology (KIAT)

\bibliographystyle{acl_natbib}
\bibliography{anthology,refs}

\begin{thebibliography}{51}
\expandafter\ifx\csname natexlab\endcsname\relax\def\natexlab#1{#1}\fi

\bibitem[{Adiwardana et~al.(2020)Adiwardana, Luong, So, Hall, Fiedel,
  Thoppilan, Yang, Kulshreshtha, Nemade, Lu et~al.}]{adiwardana2020towards}
Daniel Adiwardana, Minh-Thang Luong, David~R So, Jamie Hall, Noah Fiedel, Romal
  Thoppilan, Zi~Yang, Apoorv Kulshreshtha, Gaurav Nemade, Yifeng Lu, et~al.
  2020.
\newblock Towards a human-like open-domain chatbot.
\newblock \emph{arXiv preprint arXiv:2001.09977}.

\bibitem[{Amidei et~al.(2018)Amidei, Piwek, and
  Willis}]{amidei-etal-2018-rethinking}
Jacopo Amidei, Paul Piwek, and Alistair Willis. 2018.
\newblock \href {https://www.aclweb.org/anthology/C18-1281} {Rethinking the
  agreement in human evaluation tasks}.
\newblock In \emph{Proceedings of the 27th International Conference on
  Computational Linguistics}, pages 3318--3329, Santa Fe, New Mexico, USA.
  Association for Computational Linguistics.

\bibitem[{Amidei et~al.(2019)Amidei, Piwek, and Willis}]{amidei2019agreement}
Jacopo Amidei, Paul Piwek, and Alistair Willis. 2019.
\newblock Agreement is overrated: A plea for correlation to assess human
  evaluation reliability.

\bibitem[{Baheti et~al.(2018)Baheti, Ritter, Li, and
  Dolan}]{baheti2018generating}
Ashutosh Baheti, Alan Ritter, Jiwei Li, and Bill Dolan. 2018.
\newblock \href {https://doi.org/10.18653/v1/D18-1431} {Generating more
  interesting responses in neural conversation models with distributional
  constraints}.
\newblock In \emph{Proceedings of the 2018 Conference on Empirical Methods in
  Natural Language Processing}, pages 3970--3980, Brussels, Belgium.
  Association for Computational Linguistics.

\bibitem[{Baumgartner et~al.(2020)Baumgartner, Zannettou, Keegan, Squire, and
  Blackburn}]{baumgartner2020pushshift}
Jason Baumgartner, Savvas Zannettou, Brian Keegan, Megan Squire, and Jeremy
  Blackburn. 2020.
\newblock The pushshift reddit dataset.
\newblock In \emph{Proceedings of the International AAAI Conference on Web and
  Social Media}, volume~14, pages 830--839.

\bibitem[{Berg-Kirkpatrick et~al.(2012)Berg-Kirkpatrick, Burkett, and
  Klein}]{berg2012empirical}
Taylor Berg-Kirkpatrick, David Burkett, and Dan Klein. 2012.
\newblock An empirical investigation of statistical significance in nlp.
\newblock In \emph{Proceedings of the 2012 Joint Conference on Empirical
  Methods in Natural Language Processing and Computational Natural Language
  Learning}, pages 995--1005.

\bibitem[{Bojar(2018)}]{bojar2018english}
Ond{\v{r}}ej Bojar. 2018.
\newblock English-to-czech mt: Large data and beyond.

\bibitem[{Bradley and Terry(1952)}]{bradley1952rank}
Ralph~Allan Bradley and Milton~E Terry. 1952.
\newblock Rank analysis of incomplete block designs: I. the method of paired
  comparisons.
\newblock \emph{Biometrika}, 39(3/4):324--345.

\bibitem[{Brown et~al.(2020)Brown, Mann, Ryder, Subbiah, Kaplan, Dhariwal,
  Neelakantan, Shyam, Sastry, Askell et~al.}]{brown2020language}
Tom~B Brown, Benjamin Mann, Nick Ryder, Melanie Subbiah, Jared Kaplan, Prafulla
  Dhariwal, Arvind Neelakantan, Pranav Shyam, Girish Sastry, Amanda Askell,
  et~al. 2020.
\newblock Language models are few-shot learners.
\newblock \emph{arXiv preprint arXiv:2005.14165}.

\bibitem[{Burtsev et~al.(2018)Burtsev, Logacheva, Malykh, Serban, Lowe,
  Prabhumoye, Black, Rudnicky, and Bengio}]{burtsev2018first}
Mikhail Burtsev, Varvara Logacheva, Valentin Malykh, Iulian~Vlad Serban, Ryan
  Lowe, Shrimai Prabhumoye, Alan~W Black, Alexander Rudnicky, and Yoshua
  Bengio. 2018.
\newblock The first conversational intelligence challenge.
\newblock In \emph{The NIPS'17 Competition: Building Intelligent Systems},
  pages 25--46. Springer.

\bibitem[{Csaky(2019)}]{csaky2019deep}
Richard Csaky. 2019.
\newblock Deep learning based chatbot models.
\newblock \emph{arXiv preprint arXiv:1908.08835}.

\bibitem[{Cs{\'a}ky et~al.(2019)Cs{\'a}ky, Purgai, and
  Recski}]{csaky-etal-2019-improving}
Rich{\'a}rd Cs{\'a}ky, Patrik Purgai, and G{\'a}bor Recski. 2019.
\newblock \href {https://doi.org/10.18653/v1/P19-1567} {Improving neural
  conversational models with entropy-based data filtering}.
\newblock In \emph{Proceedings of the 57th Annual Meeting of the Association
  for Computational Linguistics}, pages 5650--5669, Florence, Italy.
  Association for Computational Linguistics.

\bibitem[{Danescu-Niculescu-Mizil and
  Lee(2011)}]{Danescu-Niculescu-Mizil+Lee:11a}
Cristian Danescu-Niculescu-Mizil and Lillian Lee. 2011.
\newblock Chameleons in imagined conversations: A new approach to understanding
  coordination of linguistic style in dialogs.
\newblock In \emph{Proceedings of the Workshop on Cognitive Modeling and
  Computational Linguistics, ACL 2011}.

\bibitem[{DeVault et~al.(2011)DeVault, Leuski, and Sagae}]{devault2011toward}
David DeVault, Anton Leuski, and Kenji Sagae. 2011.
\newblock Toward learning and evaluation of dialogue policies with text
  examples.
\newblock In \emph{Proceedings of the SIGDIAL 2011 Conference}, pages 39--48.
  Association for Computational Linguistics.

\bibitem[{Dinan et~al.(2019)Dinan, Logacheva, Malykh, Miller, Shuster, Urbanek,
  Kiela, Szlam, Serban, Lowe et~al.}]{dinan2019second}
Emily Dinan, Varvara Logacheva, Valentin Malykh, Alexander Miller, Kurt
  Shuster, Jack Urbanek, Douwe Kiela, Arthur Szlam, Iulian Serban, Ryan Lowe,
  et~al. 2019.
\newblock The second conversational intelligence challenge (convai2).
\newblock \emph{arXiv preprint arXiv:1902.00098}.

\bibitem[{Dinan et~al.(2018)Dinan, Roller, Shuster, Fan, Auli, and
  Weston}]{dinan2018wizard}
Emily Dinan, Stephen Roller, Kurt Shuster, Angela Fan, Michael Auli, and Jason
  Weston. 2018.
\newblock Wizard of wikipedia: Knowledge-powered conversational agents.
\newblock \emph{arXiv preprint arXiv:1811.01241}.

\bibitem[{Dror et~al.(2018)Dror, Baumer, Shlomov, and
  Reichart}]{dror2018hitchhiker}
Rotem Dror, Gili Baumer, Segev Shlomov, and Roi Reichart. 2018.
\newblock The hitchhiker’s guide to testing statistical significance in
  natural language processing.
\newblock In \emph{Proceedings of the 56th Annual Meeting of the Association
  for Computational Linguistics (Volume 1: Long Papers)}, pages 1383--1392.

\bibitem[{D’Haro et~al.(2020)D’Haro, Yoshino, Hori, Marks, Polymenakos,
  Kummerfeld, Galley, and Gao}]{d2020overview}
Luis~Fernando D’Haro, Koichiro Yoshino, Chiori Hori, Tim~K Marks, Lazaros
  Polymenakos, Jonathan~K Kummerfeld, Michel Galley, and Xiang Gao. 2020.
\newblock Overview of the seventh dialog system technology challenge: Dstc7.
\newblock \emph{Computer Speech \& Language}, page 101068.

\bibitem[{Fleiss(1971)}]{fleiss1971measuring}
Joseph~L Fleiss. 1971.
\newblock Measuring nominal scale agreement among many raters.
\newblock \emph{Psychological bulletin}, 76(5):378.

\bibitem[{Gao et~al.(2019)Gao, Galley, and Li}]{gao2019neural}
Jianfeng Gao, Michel Galley, and Lihong Li. 2019.
\newblock \emph{Neural Approaches to Conversational AI: Question Answering,
  Task-oriented Dialogues and Social Chatbots}.
\newblock Now Foundations and Trends.

\bibitem[{Henrysson(1963)}]{henrysson1963correction}
Sten Henrysson. 1963.
\newblock Correction of item-total correlations in item analysis.
\newblock \emph{Psychometrika}, 28(2):211--218.

\bibitem[{Herbrich et~al.(2007)Herbrich, Minka, and
  Graepel}]{herbrich2007trueskill}
Ralf Herbrich, Tom Minka, and Thore Graepel. 2007.
\newblock Trueskill™: a bayesian skill rating system.
\newblock In \emph{Advances in neural information processing systems}, pages
  569--576.

\bibitem[{Higashinaka et~al.(2017)Higashinaka, Funakoshi, Inaba, Tsunomori,
  Takahashi, and Kaji}]{higashinaka2017overview}
Ryuichiro Higashinaka, Kotaro Funakoshi, Michimasa Inaba, Yuiko Tsunomori,
  Tetsuro Takahashi, and Nobuhiro Kaji. 2017.
\newblock Overview of dialogue breakdown detection challenge 3.
\newblock \emph{Proceedings of dialog system technology challenge}, 6.

\bibitem[{Higashinaka et~al.(2016)Higashinaka, Funakoshi, Kobayashi, and
  Inaba}]{higashinaka2016dialogue}
Ryuichiro Higashinaka, Kotaro Funakoshi, Yuka Kobayashi, and Michimasa Inaba.
  2016.
\newblock The dialogue breakdown detection challenge: Task description,
  datasets, and evaluation metrics.
\newblock In \emph{Proceedings of the Tenth International Conference on
  Language Resources and Evaluation (LREC'16)}, pages 3146--3150.

\bibitem[{Klein et~al.(2017)Klein, Kim, Deng, Senellart, and Rush}]{opennmt}
Guillaume Klein, Yoon Kim, Yuntian Deng, Jean Senellart, and Alexander~M. Rush.
  2017.
\newblock \href {https://doi.org/10.18653/v1/P17-4012} {Open{NMT}: Open-source
  toolkit for neural machine translation}.
\newblock In \emph{Proc. ACL}.

\bibitem[{Li et~al.(2016{\natexlab{a}})Li, Galley, Brockett, Gao, and
  Dolan}]{li2015diversity}
Jiwei Li, Michel Galley, Chris Brockett, Jianfeng Gao, and Bill Dolan.
  2016{\natexlab{a}}.
\newblock \href {https://doi.org/10.18653/v1/N16-1014} {A diversity-promoting
  objective function for neural conversation models}.
\newblock In \emph{Proceedings of the 2016 Conference of the North {A}merican
  Chapter of the Association for Computational Linguistics: Human Language
  Technologies}, pages 110--119, San Diego, California. Association for
  Computational Linguistics.

\bibitem[{Li et~al.(2016{\natexlab{b}})Li, Galley, Brockett, Spithourakis, Gao,
  and Dolan}]{li2016persona}
Jiwei Li, Michel Galley, Chris Brockett, Georgios Spithourakis, Jianfeng Gao,
  and Bill Dolan. 2016{\natexlab{b}}.
\newblock \href {https://doi.org/10.18653/v1/P16-1094} {A persona-based neural
  conversation model}.
\newblock In \emph{Proceedings of the 54th Annual Meeting of the Association
  for Computational Linguistics (Volume 1: Long Papers)}, pages 994--1003,
  Berlin, Germany. Association for Computational Linguistics.

\bibitem[{Li et~al.(2017{\natexlab{a}})Li, Monroe, Shi, Jean, Ritter, and
  Jurafsky}]{li2017adversarial}
Jiwei Li, Will Monroe, Tianlin Shi, S{\'e}bastien Jean, Alan Ritter, and Dan
  Jurafsky. 2017{\natexlab{a}}.
\newblock \href {https://doi.org/10.18653/v1/D17-1230} {Adversarial learning
  for neural dialogue generation}.
\newblock In \emph{Proceedings of the 2017 Conference on Empirical Methods in
  Natural Language Processing}, pages 2157--2169, Copenhagen, Denmark.
  Association for Computational Linguistics.

\bibitem[{Li et~al.(2019)Li, Weston, and Roller}]{li2019acute}
Margaret Li, Jason Weston, and Stephen Roller. 2019.
\newblock Acute-eval: Improved dialogue evaluation with optimized questions and
  multi-turn comparisons.
\newblock \emph{arXiv preprint arXiv:1909.03087}.

\bibitem[{Li et~al.(2017{\natexlab{b}})Li, Su, Shen, Li, Cao, and
  Niu}]{li-etal-2017-dailydialog}
Yanran Li, Hui Su, Xiaoyu Shen, Wenjie Li, Ziqiang Cao, and Shuzi Niu.
  2017{\natexlab{b}}.
\newblock \href {https://www.aclweb.org/anthology/I17-1099} {{D}aily{D}ialog: A
  manually labelled multi-turn dialogue dataset}.
\newblock In \emph{Proceedings of the Eighth International Joint Conference on
  Natural Language Processing (Volume 1: Long Papers)}, pages 986--995, Taipei,
  Taiwan. Asian Federation of Natural Language Processing.

\bibitem[{Liu et~al.(2016)Liu, Lowe, Serban, Noseworthy, Charlin, and
  Pineau}]{liu2016not}
Chia-Wei Liu, Ryan Lowe, Iulian Serban, Mike Noseworthy, Laurent Charlin, and
  Joelle Pineau. 2016.
\newblock \href {https://doi.org/10.18653/v1/D16-1230} {How {NOT} to evaluate
  your dialogue system: An empirical study of unsupervised evaluation metrics
  for dialogue response generation}.
\newblock In \emph{Proceedings of the 2016 Conference on Empirical Methods in
  Natural Language Processing}, pages 2122--2132, Austin, Texas. Association
  for Computational Linguistics.

\bibitem[{Logacheva et~al.(2018)Logacheva, Burtsev, Malykh, Poluliakh,
  Rudnicky, Serban, Lowe, Prabhumoye, Black, and Bengio}]{logacheva2018dataset}
Varvara Logacheva, Mikhail Burtsev, Valentin Malykh, Vadim Poluliakh, Alexander
  Rudnicky, Iulian Serban, Ryan Lowe, Shrimai Prabhumoye, Alan~W Black, and
  Yoshua Bengio. 2018.
\newblock A dataset of topic-oriented human-to-chatbot dialogues.

\bibitem[{Lowe et~al.(2017)Lowe, Noseworthy, Serban, Angelard-Gontier, Bengio,
  and Pineau}]{lowe2017towards}
Ryan Lowe, Michael Noseworthy, Iulian~V Serban, Nicolas Angelard-Gontier,
  Yoshua Bengio, and Joelle Pineau. 2017.
\newblock Towards an automatic turing test: Learning to evaluate dialogue
  responses.
\newblock \emph{arXiv preprint arXiv:1708.07149}.

\bibitem[{Miller et~al.(2017)Miller, Feng, Batra, Bordes, Fisch, Lu, Parikh,
  and Weston}]{miller2017parlai}
Alexander Miller, Will Feng, Dhruv Batra, Antoine Bordes, Adam Fisch, Jiasen
  Lu, Devi Parikh, and Jason Weston. 2017.
\newblock \href {https://doi.org/10.18653/v1/D17-2014} {{P}arl{AI}: A dialog
  research software platform}.
\newblock In \emph{Proceedings of the 2017 Conference on Empirical Methods in
  Natural Language Processing: System Demonstrations}, pages 79--84,
  Copenhagen, Denmark. Association for Computational Linguistics.

\bibitem[{Novikova et~al.(2017)Novikova, Du{\v{s}}ek, Cercas~Curry, and
  Rieser}]{novikova2017we}
Jekaterina Novikova, Ond{\v{r}}ej Du{\v{s}}ek, Amanda Cercas~Curry, and Verena
  Rieser. 2017.
\newblock \href {https://doi.org/10.18653/v1/D17-1238} {Why we need new
  evaluation metrics for {NLG}}.
\newblock In \emph{Proceedings of the 2017 Conference on Empirical Methods in
  Natural Language Processing}, pages 2241--2252, Copenhagen, Denmark.
  Association for Computational Linguistics.

\bibitem[{Novikova et~al.(2018)Novikova, Du{\v{s}}ek, and
  Rieser}]{Novikova2018}
Jekaterina Novikova, Ondrej Du{\v{s}}ek, and Verena Rieser. 2018.
\newblock Rank{ME}: Reliable human ratings for natural language generation.
\newblock In \emph{NAACL}, New Orleans, Louisiana.

\bibitem[{Papineni et~al.(2002)Papineni, Roukos, Ward, and
  Zhu}]{papineni2002bleu}
Kishore Papineni, Salim Roukos, Todd Ward, and Wei-Jing Zhu. 2002.
\newblock Bleu: a method for automatic evaluation of machine translation.
\newblock In \emph{Proceedings of the 40th annual meeting of the Association
  for Computational Linguistics}, pages 311--318.

\bibitem[{Pennington et~al.(2014)Pennington, Socher, and
  Manning}]{pennington2014glove}
Jeffrey Pennington, Richard Socher, and Christopher~D Manning. 2014.
\newblock Glove: Global vectors for word representation.
\newblock In \emph{Proceedings of the 2014 conference on empirical methods in
  natural language processing (EMNLP)}, pages 1532--1543.

\bibitem[{Radford et~al.(2019)Radford, Wu, Child, Luan, Amodei, and
  Sutskever}]{radford2019language}
Alec Radford, Jeffrey Wu, Rewon Child, David Luan, Dario Amodei, and Ilya
  Sutskever. 2019.
\newblock Language models are unsupervised multitask learners.
\newblock \emph{OpenAI Blog}, 1(8):9.

\bibitem[{Rashkin et~al.(2019)Rashkin, Smith, Li, and
  Boureau}]{rashkin2018towards}
Hannah Rashkin, Eric~Michael Smith, Margaret Li, and Y-Lan Boureau. 2019.
\newblock Towards empathetic open-domain conversation models: A new benchmark
  and dataset.
\newblock In \emph{Proceedings of the 57th Annual Meeting of the Association
  for Computational Linguistics}, pages 5370--5381.

\bibitem[{Roller et~al.(2020)Roller, Dinan, Goyal, Ju, Williamson, Liu, Xu,
  Ott, Shuster, Smith et~al.}]{roller2020recipes}
Stephen Roller, Emily Dinan, Naman Goyal, Da~Ju, Mary Williamson, Yinhan Liu,
  Jing Xu, Myle Ott, Kurt Shuster, Eric~M Smith, et~al. 2020.
\newblock Recipes for building an open-domain chatbot.
\newblock \emph{arXiv preprint arXiv:2004.13637}.

\bibitem[{Sakaguchi et~al.(2014)Sakaguchi, Post, and
  Van~Durme}]{sakaguchi2014efficient}
Keisuke Sakaguchi, Matt Post, and Benjamin Van~Durme. 2014.
\newblock Efficient elicitation of annotations for human evaluation of machine
  translation.
\newblock In \emph{Proceedings of the Ninth Workshop on Statistical Machine
  Translation}, pages 1--11.

\bibitem[{Sedoc et~al.(2019)Sedoc, Ippolito, Kirubarajan, Thirani, Ungar, and
  Callison-Burch}]{sedoc-etal-2019-chateval}
Jo{\~a}o Sedoc, Daphne Ippolito, Arun Kirubarajan, Jai Thirani, Lyle Ungar, and
  Chris Callison-Burch. 2019.
\newblock \href {https://doi.org/10.18653/v1/N19-4011} {{C}hat{E}val: A tool
  for chatbot evaluation}.
\newblock In \emph{Proceedings of the 2019 Conference of the North {A}merican
  Chapter of the Association for Computational Linguistics (Demonstrations)},
  pages 60--65, Minneapolis, Minnesota. Association for Computational
  Linguistics.

\bibitem[{See et~al.(2019)See, Roller, Kiela, and Weston}]{see2019makes}
Abigail See, Stephen Roller, Douwe Kiela, and Jason Weston. 2019.
\newblock What makes a good conversation? how controllable attributes affect
  human judgments.
\newblock In \emph{Proceedings of the 2019 Conference of the North American
  Chapter of the Association for Computational Linguistics: Human Language
  Technologies, Volume 1 (Long and Short Papers)}, pages 1702--1723.

\bibitem[{Shuster et~al.(2019)Shuster, Ju, Roller, Dinan, Boureau, and
  Weston}]{shuster2019dialogue}
Kurt Shuster, Da~Ju, Stephen Roller, Emily Dinan, Y-Lan Boureau, and Jason
  Weston. 2019.
\newblock The dialogue dodecathlon: Open-domain knowledge and image grounded
  conversational agents.
\newblock \emph{arXiv preprint arXiv:1911.03768}.

\bibitem[{Smith et~al.(2020)Smith, Williamson, Shuster, Weston, and
  Boureau}]{smith2020can}
Eric~Michael Smith, Mary Williamson, Kurt Shuster, Jason Weston, and Y-Lan
  Boureau. 2020.
\newblock Can you put it all together: Evaluating conversational agents'
  ability to blend skills.
\newblock \emph{arXiv preprint arXiv:2004.08449}.

\bibitem[{Vaswani et~al.(2017)Vaswani, Shazeer, Parmar, Uszkoreit, Jones,
  Gomez, Kaiser, and Polosukhin}]{vaswani2017attention}
Ashish Vaswani, Noam Shazeer, Niki Parmar, Jakob Uszkoreit, Llion Jones,
  Aidan~N Gomez, {\L}ukasz Kaiser, and Illia Polosukhin. 2017.
\newblock Attention is all you need.
\newblock In \emph{Advances in neural information processing systems}, pages
  5998--6008.

\bibitem[{Venkatesh et~al.(2018)Venkatesh, Khatri, Ram, Guo, Gabriel, Nagar,
  Prasad, Cheng, Hedayatnia, Metallinou, Goel, Yang, and Raju}]{Venkatesh2018}
Anu Venkatesh, Chandra Khatri, Ashwin Ram, Fenfei Guo, Raefer Gabriel, Ashish
  Nagar, Rohit Prasad, Ming Cheng, Behnam Hedayatnia, Angeliki Metallinou,
  Rahul Goel, Shaohua Yang, and Anirudh Raju. 2018.
\newblock \href {http://arxiv.org/abs/1801.03625} {{On Evaluating and Comparing
  Conversational Agents}}.
\newblock (Nips):1--10.

\bibitem[{Vinyals and Le(2015)}]{vinyals2015neural}
Oriol Vinyals and Quoc Le. 2015.
\newblock A neural conversational model.
\newblock \emph{arXiv preprint arXiv:1506.05869}.

\bibitem[{Zhang et~al.(2018)Zhang, Dinan, Urbanek, Szlam, Kiela, and
  Weston}]{zhang2018personalizing}
Saizheng Zhang, Emily Dinan, Jack Urbanek, Arthur Szlam, Douwe Kiela, and Jason
  Weston. 2018.
\newblock Personalizing dialogue agents: I have a dog, do you have pets too?
\newblock In \emph{Proceedings of the 56th Annual Meeting of the Association
  for Computational Linguistics (Volume 1: Long Papers)}, pages 2204--2213.

\bibitem[{Zhang et~al.(2019)Zhang, Sun, Galley, Chen, Brockett, Gao, Gao, Liu,
  and Dolan}]{zhang2019dialogpt}
Yizhe Zhang, Siqi Sun, Michel Galley, Yen-Chun Chen, Chris Brockett, Xiang Gao,
  Jianfeng Gao, Jingjing Liu, and Bill Dolan. 2019.
\newblock Dialogpt: Large-scale generative pre-training for conversational
  response generation.
\newblock \emph{arXiv preprint arXiv:1911.00536}.

\end{thebibliography}

\clearpage

\appendix

\section{ChatEval Amazon Mechanical Turk Interface}
\label{sec:app_chateval}
\begin{figure*}[tbh]
    \centering
    \includegraphics[width=\textwidth]{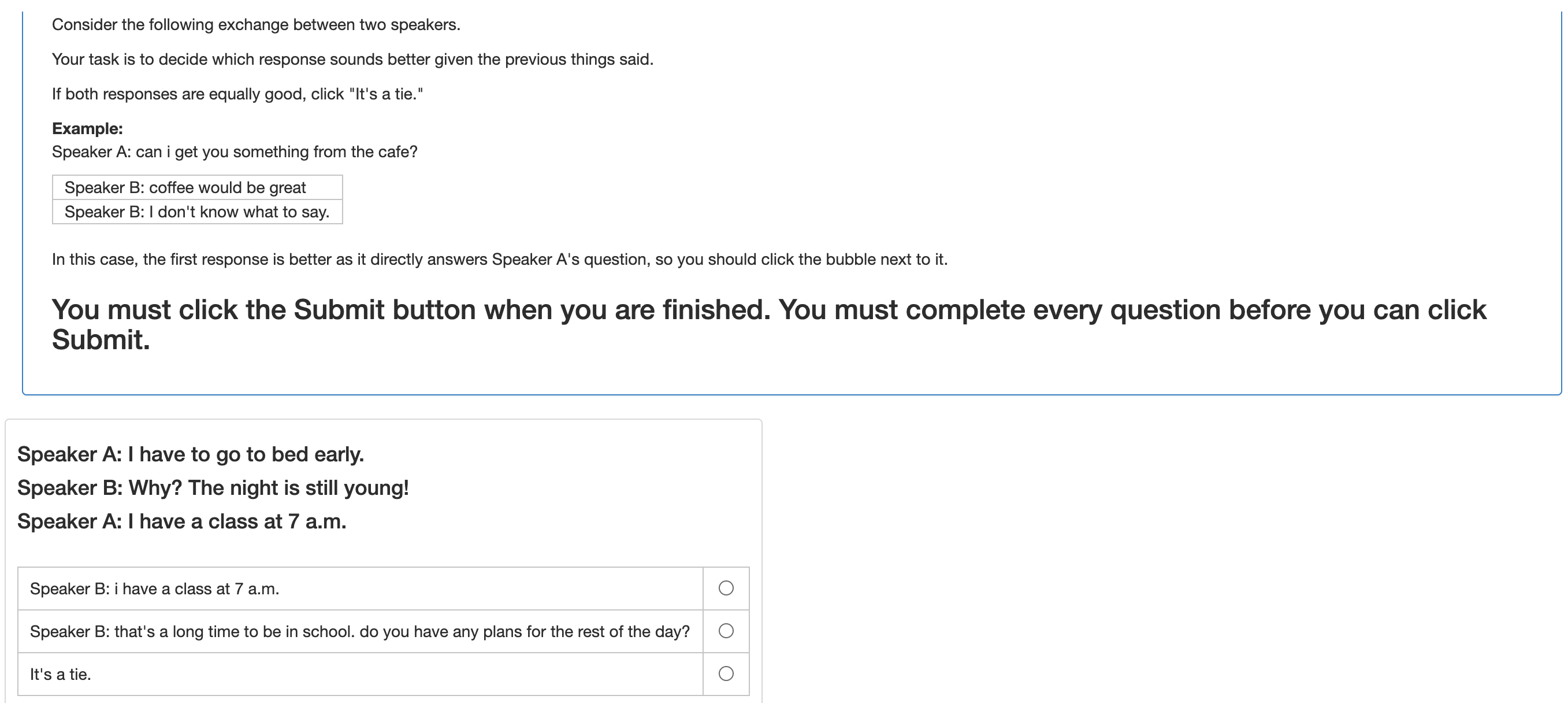}
    \caption{Screenshot of Amazon Mechanical Turk HIT.}
    \label{fig:chateval}
\end{figure*}

Figure~\ref{fig:chateval} is the Amazon Mechanical Turk (AMT) Human Intelligence Task (HIT). Each annotator is paid \$0.01  per annotations. AMT workers are shown 10 comparisons per HIT. The order in which system A vs B is presented is randomized and all output is detokenized in a standard manner. We use a minimum of three annotators. Each HIT take approximately one minute on average. A maximum of five minutes are allowed for the task in order to ensure quick completion times. On average an evaluation between two systems takes under 25 minutes.

\section{System details} \label{appen:system} 
We utilized pre-trained models from each generative system and reproduce system response code for our evaluation. We will release the deployment code soon. 

\paragraph{Blender} We used pre-trained Blender 2.7B and Blender 90M models without persona for the evaluation \footnote{\url{https://parl.ai/projects/recipes/}} in a safety interactive mode with Blended Skill Talk~\cite{smith2020can}. Blender employed a standard seq2seq transformer architecture. Blender 2.7B model used 2 encoder layers, 24 decoder layers, 32 attention heads and 2560 dimensional embeddings. Blender 90M model parameters followed by~\citet{shuster2019dialogue}. 

Blender trained with Reddit dataset~\cite{baumgartner2020pushshift} 1.5B training examples from PushShift\footnote{\url{3https://files.pushshift.io/reddit/}} through July 2019. Also, Blender fine-tuned with Blended Skill Talk, which is a mimic task such as task with ConvAI2 dataset (i.e., PersonaChat)~\cite{zhang2018personalizing}, Wizard of Wikipedia~\cite{dinan2018wizard} and Empathetic Dialogues~\cite{rashkin2018towards} for focusing personality and engaging the other speaker, empathy and knowledge.

\paragraph{Blender with persona} We used pre-trained Blender 2.7B with persona for the evaluation \footnote{\url{https://parl.ai/projects/recipes/}}. We use the persona as one of a ParlAI persona list in Table~\ref{tab:persona} followed by ParlAI document \footnote{\url{https://parl.ai/docs/tutorial_tipsntricks.html}}.

\paragraph{DialoGPT} We used pre-trained DialoGPT medium (345M) model\footnote{\url{https://github.com/microsoft/DialoGPT}} in this work. DialoGPT inherits from GPT-2~\cite{radford2019language} with 12 to 48 layer transformer. The medium model uses 24 layers. We reproduced decoder to apply the history size for the multi-turn evaluation. 

DialoGPT trained on scraped from Reddit spanning from 2005 to 2017. The dataset consists of 140 million dialogue instances.

\paragraph{ConvAI2 (seq2seq)} We used pre-trained ParlAI ConvAI2 seq2seq model \footnote{\url{https://github.com/facebookresearch/ParlAI/tree/master/projects/convai2}}. This model has LSTM architecture with GloVe~\cite{pennington2014glove} embeddings and trained on PersonaChat~\cite{zhang2018personalizing}.

\paragraph{ConvAI2 (KV-MemNN)} We used pre-trained ParlAI ConvAI2 Key-Value Profile Memory Network model \footnote{\url{https://github.com/facebookresearch/ParlAI/tree/master/projects/convai2}}. This model trained with PersonaChat data as encoding each of the profile entries into individual memory representations in a memory network. 

\paragraph{ParlAI(controllable)} We used pre-trained specificity-controlled CT model (with WD repetition control) \footnote{\url{https://parl.ai/projects/controllable_dialogue/}}, which is trained on 2.5 million Twitter message-response pairs\footnote{The Twitter dataset is provided in ParlAI; details can be
found here: \url{https://parl.ai/docs/tasks.html}} and then fine-tuned it on PersonaChat~\cite{zhang2018personalizing}. 

This model based on seq2seq model and also fine-tuned with loss\_CT as described~\citep{see2019makes}'s work. 

\paragraph{Transformer} We used pre-trained Transformer model\footnote{\url{https://github.com/ricsinaruto/Seq2seqChatbots}} trained on target-side identity clustering filtered data except on NCME, which is used with not overfitted version. The model trained on DailyDialog~\cite{li-etal-2017-dailydialog} 90K utterances in 13K dialogs. The system consists of 512 hidden size, six hidden layers and 2048 filter size. More details see~\citep{csaky-etal-2019-improving}'s work. 

\paragraph{OpenNMT(OS)} We used OpenNMT trained model with seq2seq with attention on opensubtitle questions only\footnote{\url{https://github.com/OpenNMT/OpenNMT-py}}. The model\footnote{\url{https://opennmt.net/Models-py/}} consists of 2-layer LSTM with 500 dimensional embeddings, also using global attention.

\paragraph{OpenNMT(Twitter)} We used OpenNMT trained model with seq2seq with attention trained on Twitter dataset (from ParlAI)\footnote{\url{https://github.com/facebookresearch/ParlAI/tree/master/parlai/tasks/twitter}}. 

\paragraph{CakeChat} We used pre-trained model for both single-turn and multi-turn. They trained on a preprocessed Twitter corpus with approximate 50 million dialgos (11GB of text data). They released pre-trained model on Amazon S3 for single-turn and multi-turn by running here~\footnote{\url{https://github.com/lukalabs/cakechat/blob/master/tools/fetch.py}}.

\paragraph{DC-NeuralConversation} We used DC-MMI200 model~\cite{baheti2018generating} responses from here\footnote{\url{https://github.com/abaheti95/DC-NeuralConversation/blob/master/MTurk\%20Evaluation/MTurk2\%20model\%20responses/full_model_tsim_esim_B200_MMI_decoding_cornell_mturk2_test_predictions.txt}} for evaluation. Maximum Mutual Information (MMI)~\cite{li2015diversity} was re-implemented MMI-bidi in~\citet{baheti2018generating}'s work and DC-MMI200 is MMI-bidi reranking with a beam size of 200 trained on opensubtitle dataset.

\section{Distinctness} \label{appen:violin} 
\begin{figure}[H]
\centering
    \includegraphics[width=1\linewidth]{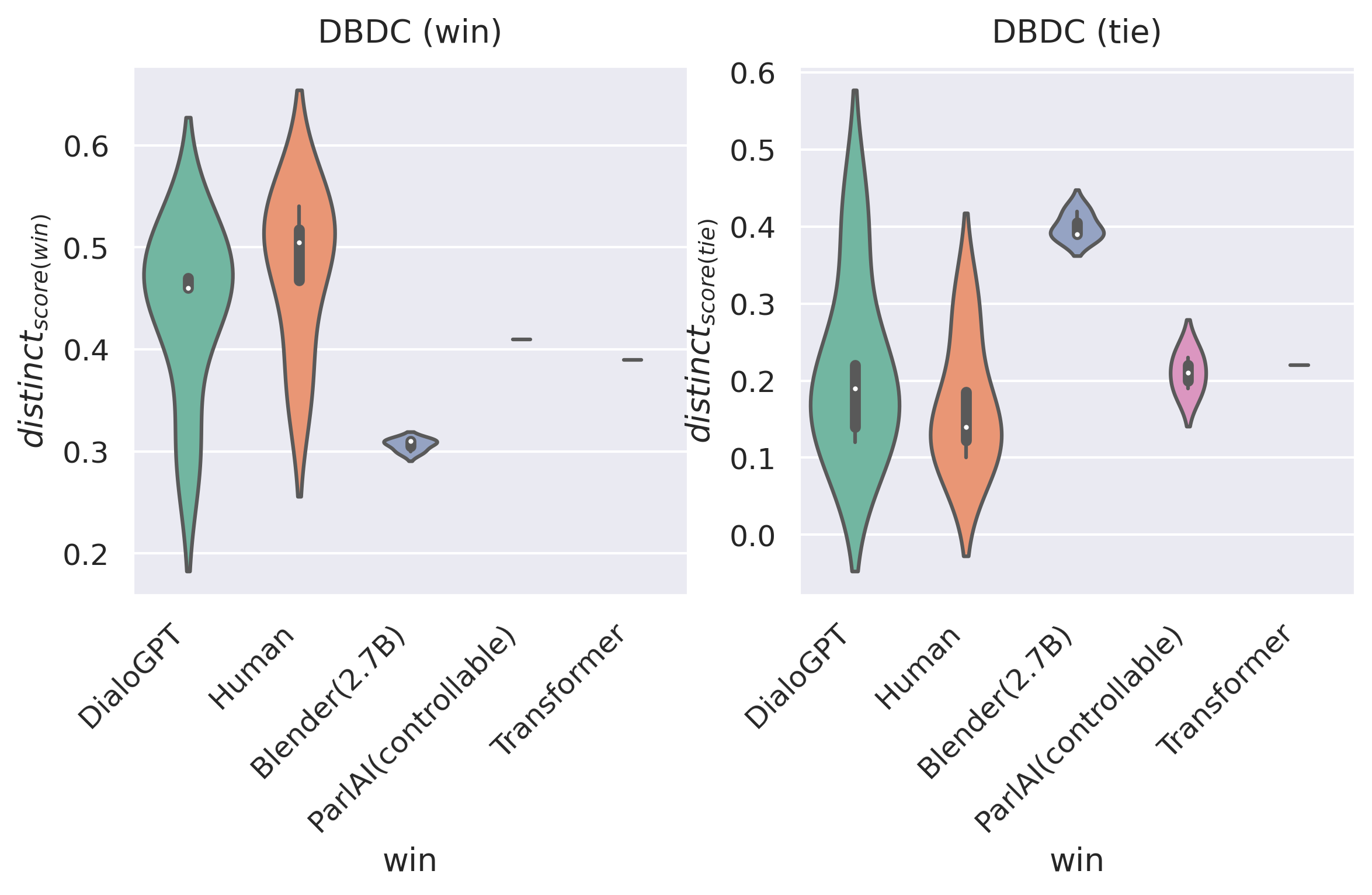}\hfil
\caption{The A/B model comparison results on DBDC. Note that the x-axis indicates wins and y-axis indicates $distinct_{score(win)}$ (left) and $distinct_{score(tie)}$ (right).}
\label{fig:dbdc_violin}
\end{figure}

\begin{figure}[H]
\centering
    \includegraphics[width=1\linewidth]{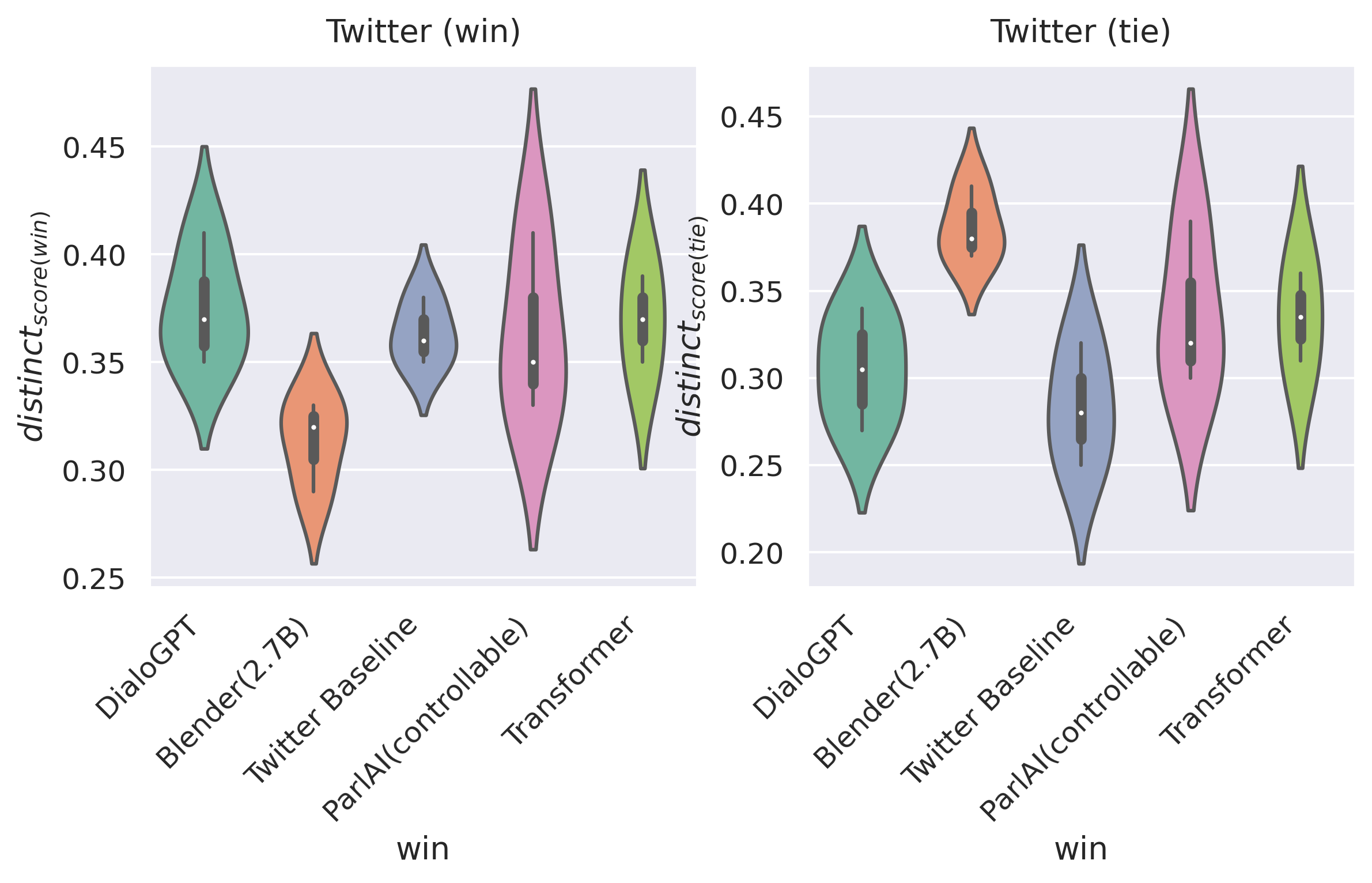}\hfil
\caption{The A/B model comparison results on Twitter.}
\label{fig:twitter_violin}
\end{figure}

\begin{figure}[H]
\centering
    \includegraphics[width=1\linewidth]{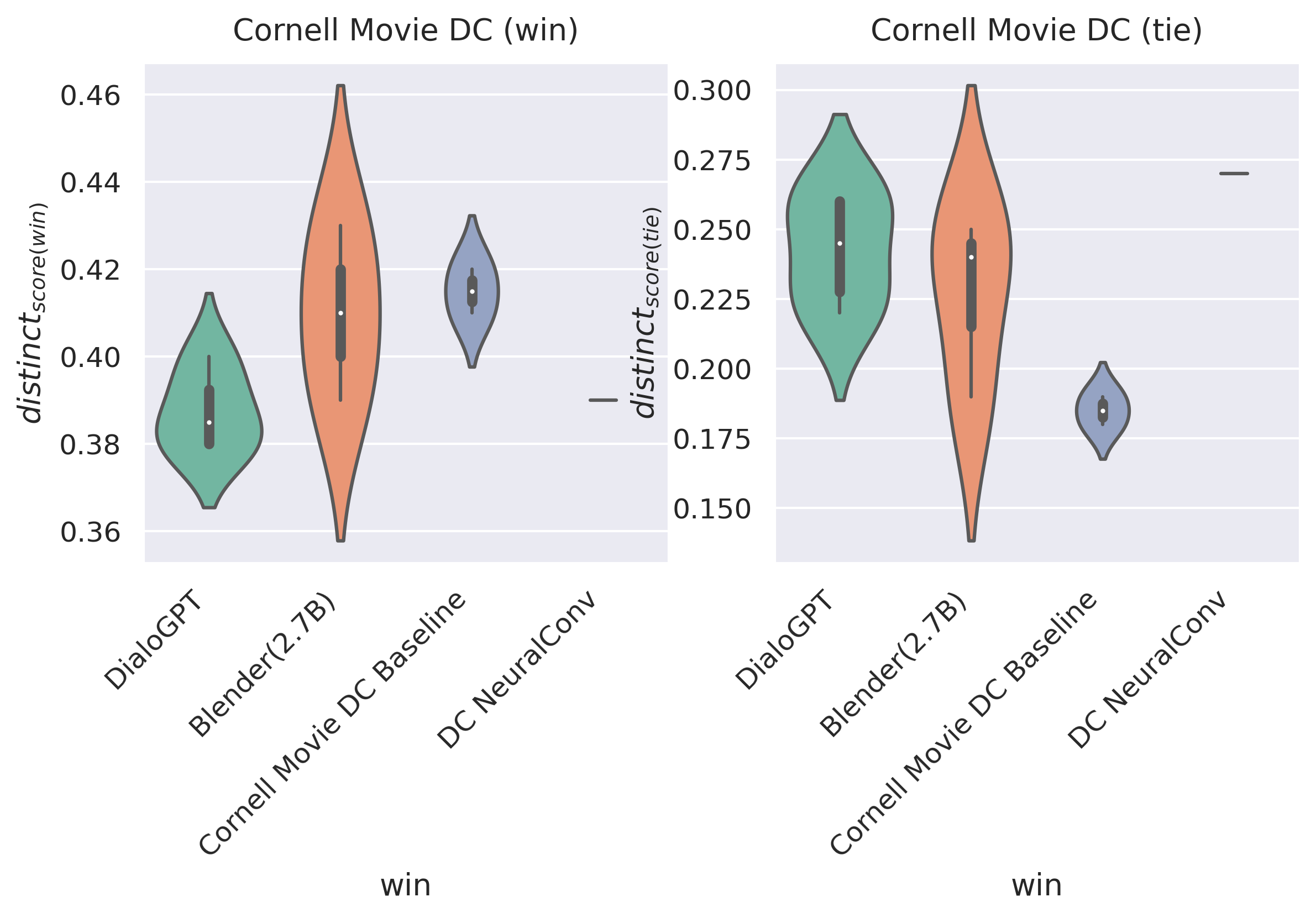}\hfil
\caption{The A/B model comparison results on Cornell Movie DC.}
\label{fig:cornell_violin}
\end{figure}

\begin{figure}[H]
\centering
    \includegraphics[width=1\linewidth]{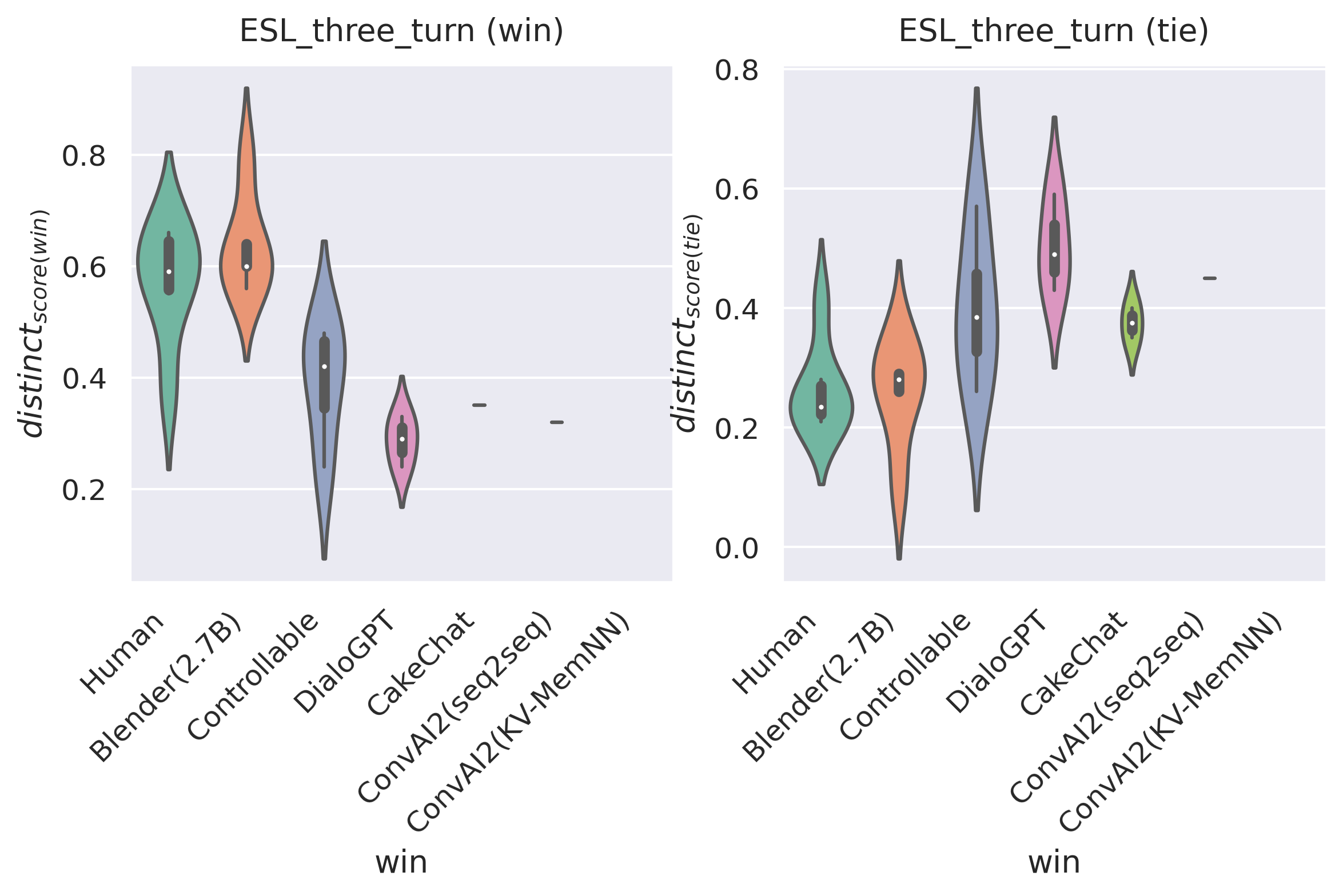}\hfil
\caption{The A/B model comparison results on ESL three turns. Noth that ConvAI2(KV-MemNN) never wins.}
\label{fig:multi_violin}
\end{figure}

\section{Prompt Validity}
\begin{figure}[tbh]
    \centering
    \includegraphics[width=0.47\textwidth]{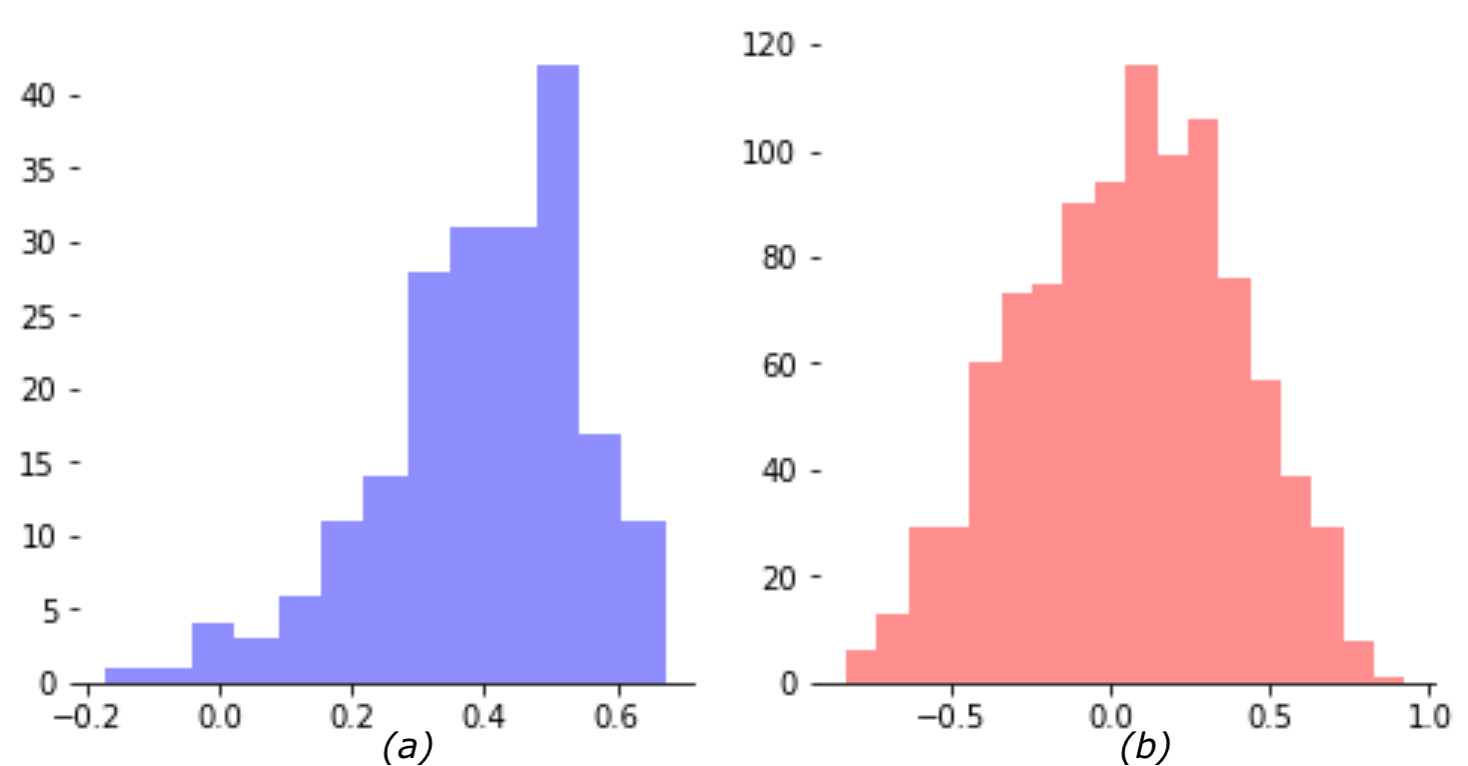}
    \caption{Correlation between prompt rating and total rating. NCME (a) and Cornell Movie D.C. (b).}
    \label{fig:prompt_validity}
\end{figure}

\section{Ranking} \label{appen:ranking} 

\begin{figure}[H]
\centering
    \includegraphics[width=0.8\linewidth]{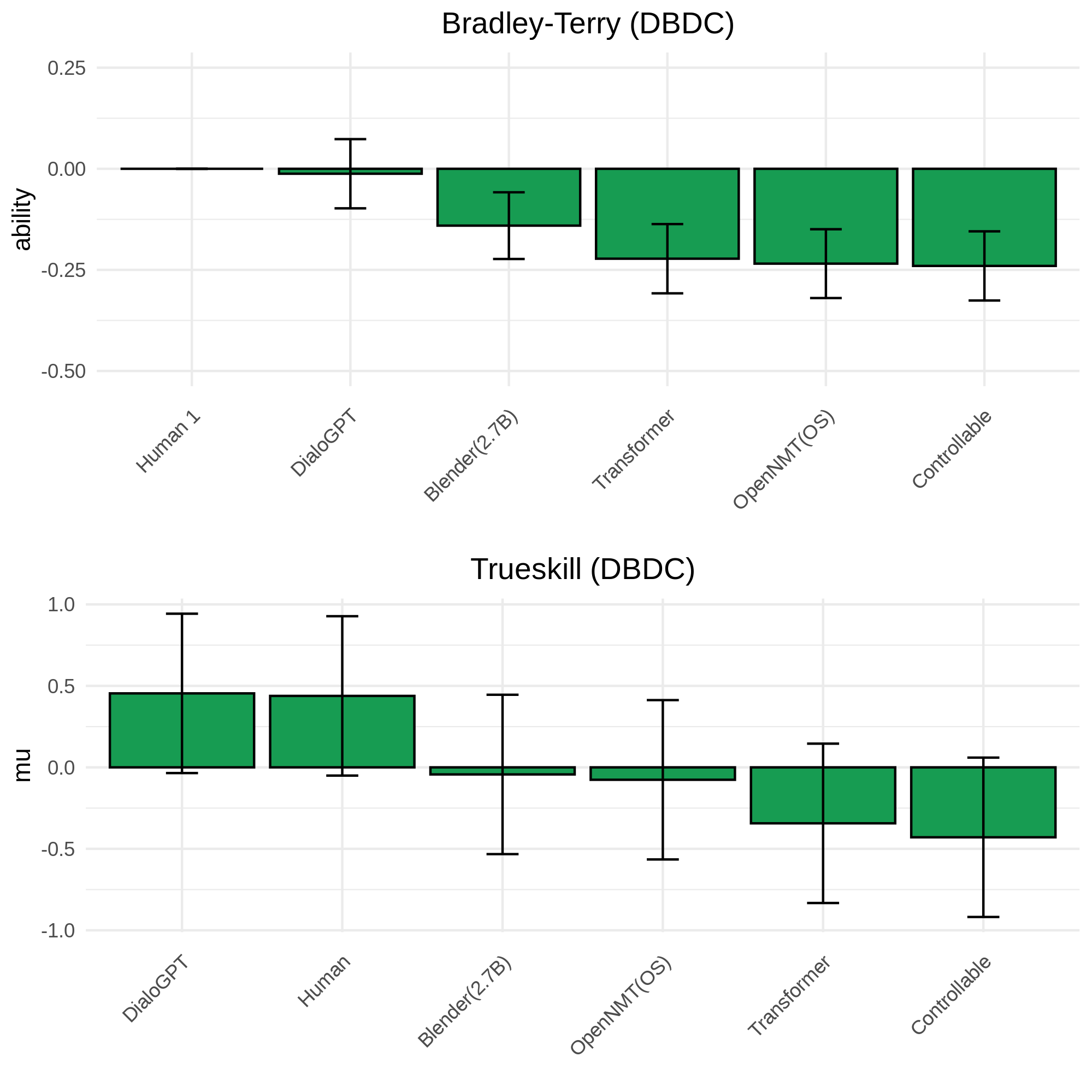}\hfil
\caption{Bradley-Terry and TrueSkill result on DBDC.}
\label{fig:dbdc_brad}
\end{figure}

\begin{figure}[H]
\centering
    \includegraphics[width=0.8\linewidth]{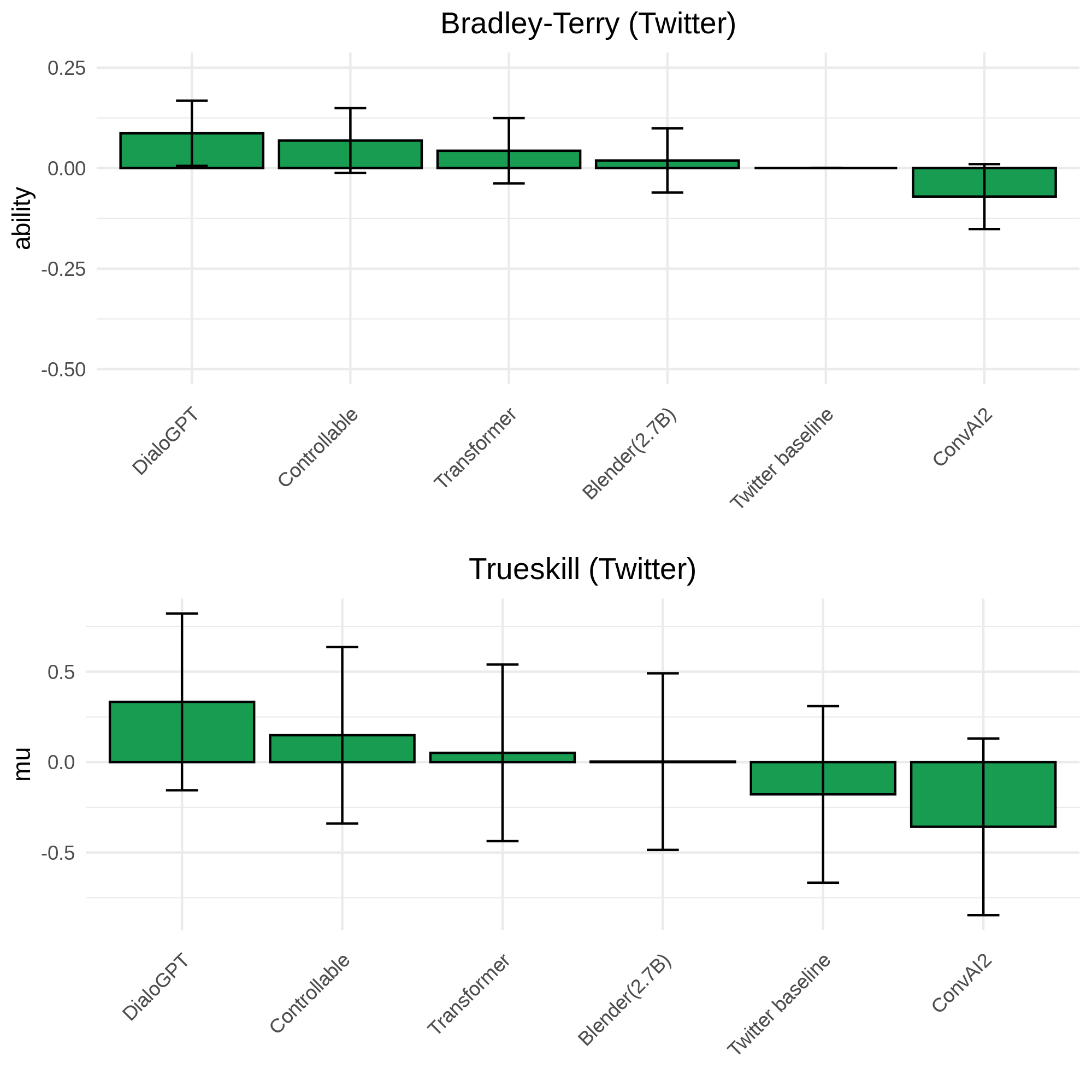}\hfil
\caption{Bradley-Terry and TrueSkill result on Twitter.}
\label{fig:twitter_brad}
\end{figure}

\begin{figure}[H]
\centering
    \includegraphics[width=0.8\linewidth]{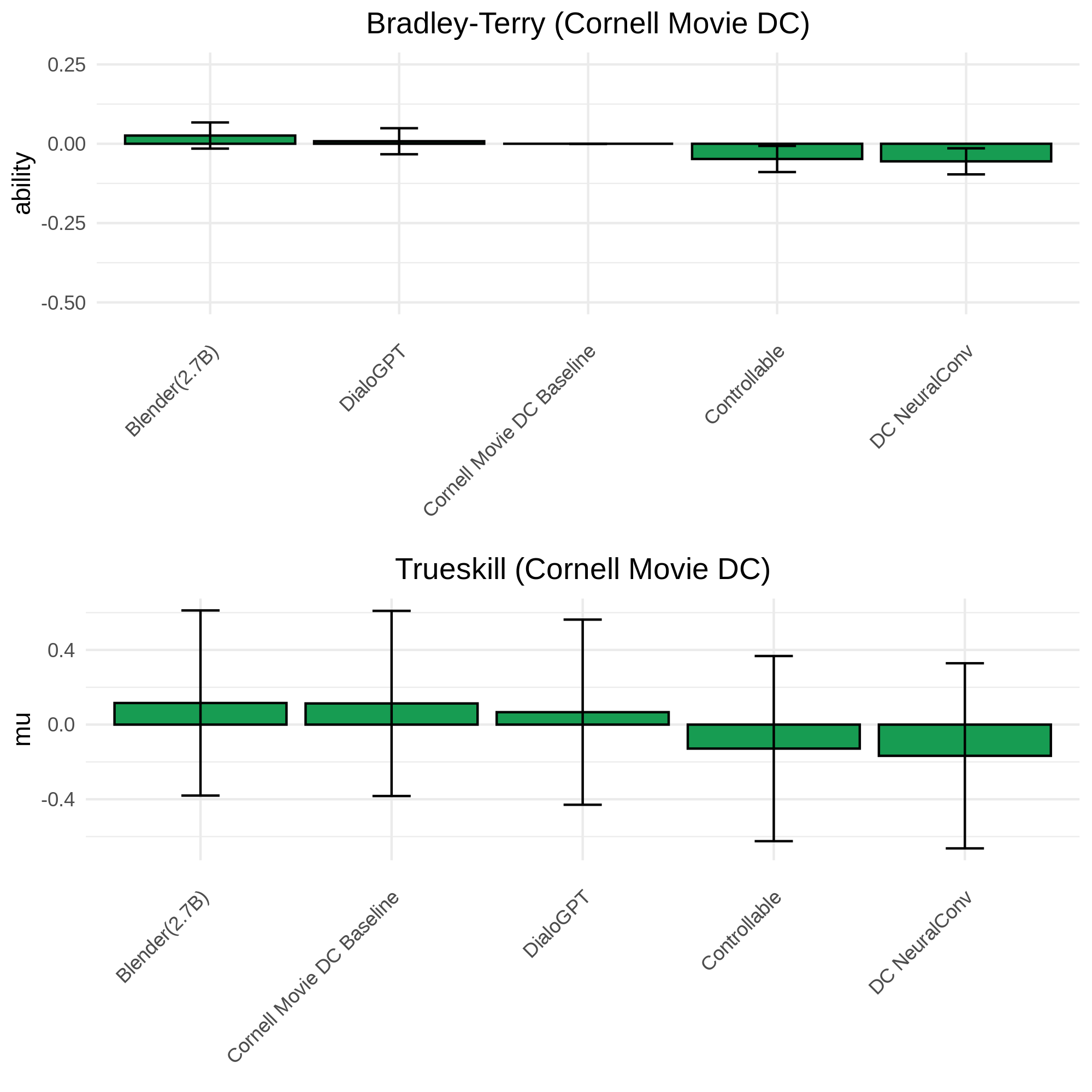}\hfil
\caption{Bradley-Terry and TrueSkill result on Cornell Movie DC.}
\label{fig:cornell_brad}
\end{figure}

\begin{figure}[tbh]
\centering
    \includegraphics[width=0.8\linewidth]{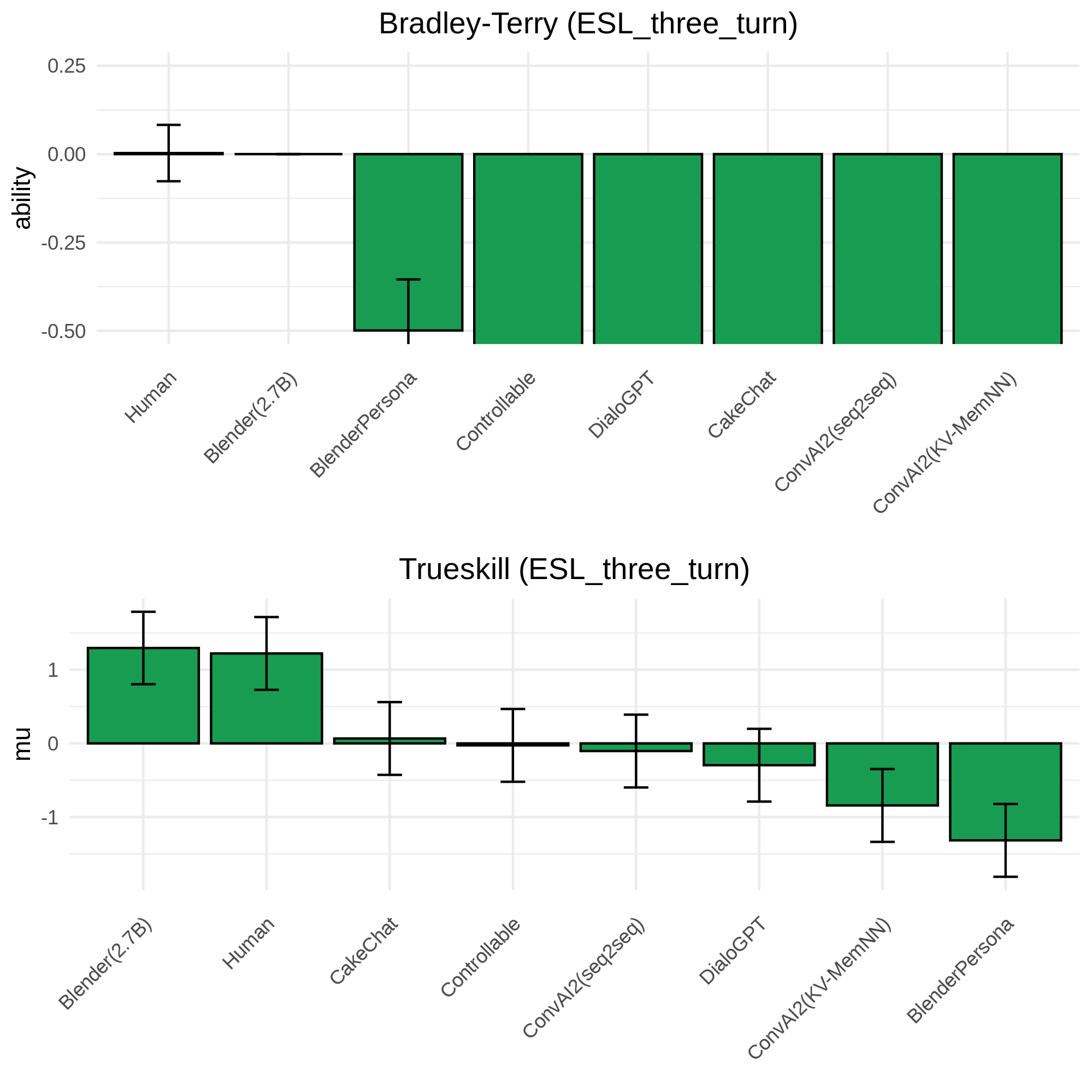}\hfil
\caption{Bradley-Terry and TrueSkill result on ESL three turns.}
\label{fig:multi_brad}
\end{figure}

\section{Annotation Correlation} \label{appen:correlation} 
We conduct spearman correlation between the agreement features and major score. Furthermore, we observe the correlation with other evaluation sets.

\begin{figure}[H]
\centering
    \includegraphics[width=1\linewidth]{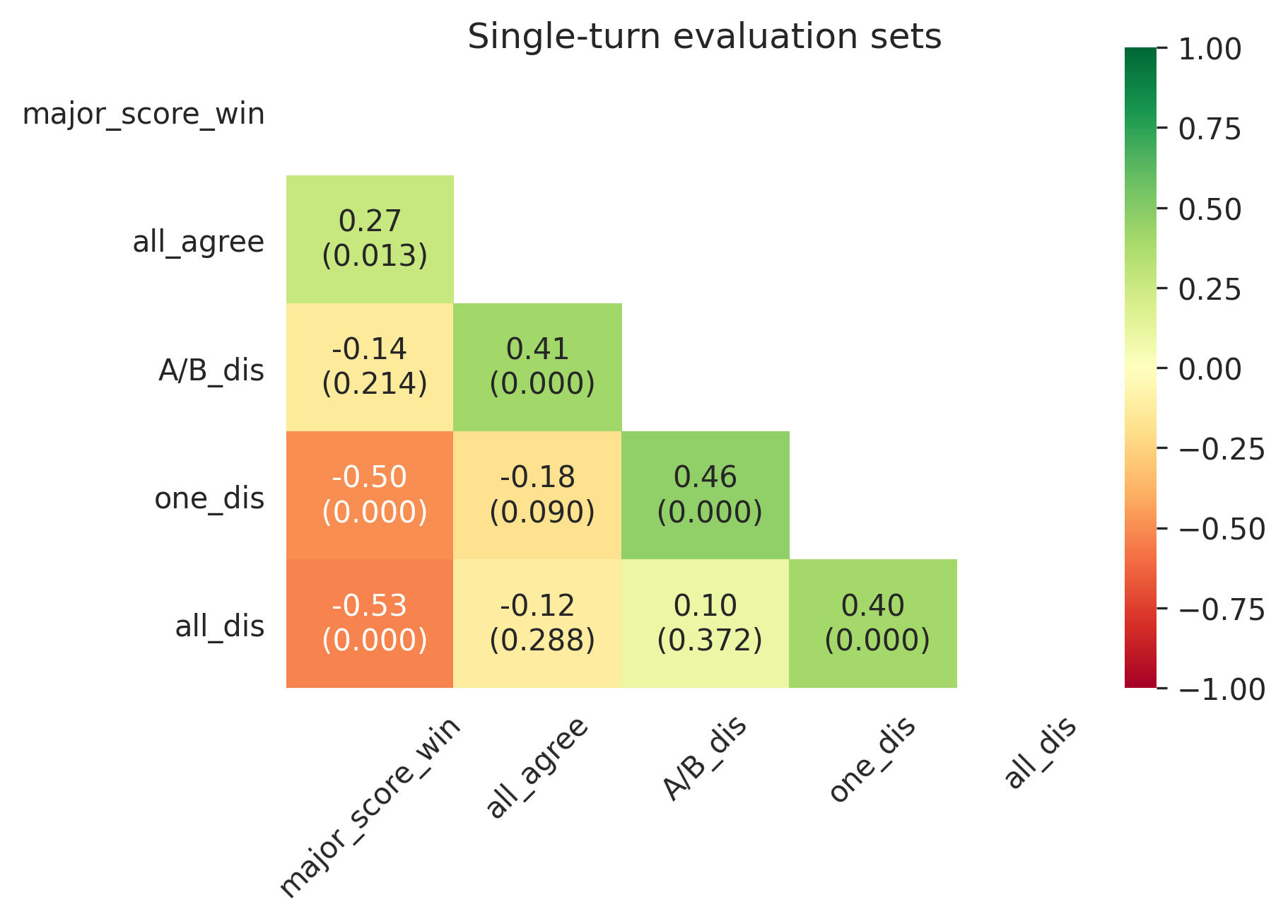}\hfil
\caption{Results on the correlation between $major_{score(win)}$ and agreement \& disagreement features in all of the single-turn evaluation sets. The weight indicates the correlation value and the value in parentheses is p-value (p-value $<$ 0.05 indicates statistically significant).}
\label{fig:aggre_corr_spear}
\end{figure}

\begin{figure}[H]
\centering
    \includegraphics[width=1\linewidth]{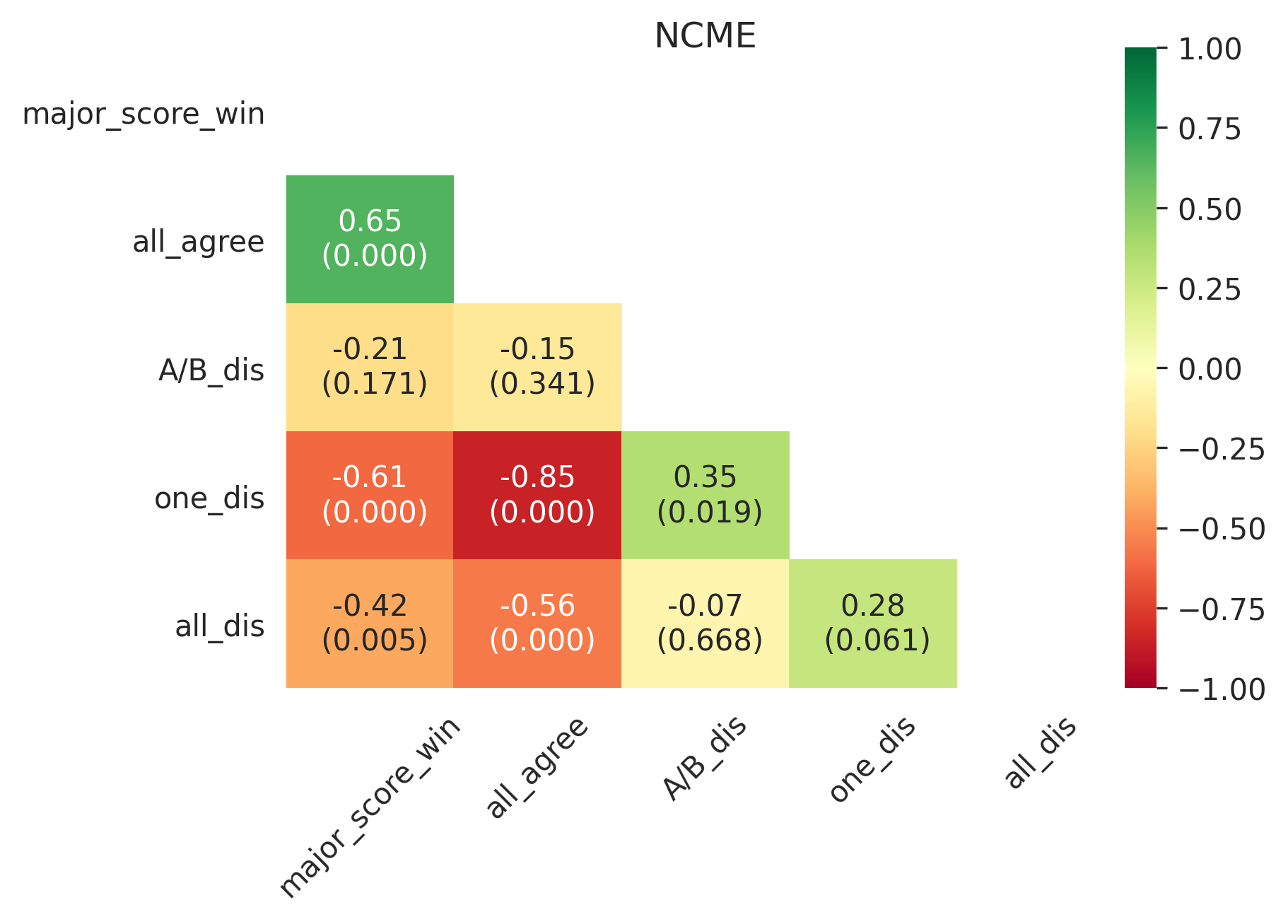}\hfil
\caption{Results on the correlation between $major_{score(win)}$ and agreement \& disagreement features in NCME.}
\label{fig:ncm_corr_spear}
\end{figure}

\begin{figure}[H]
\centering
    \includegraphics[width=1\linewidth]{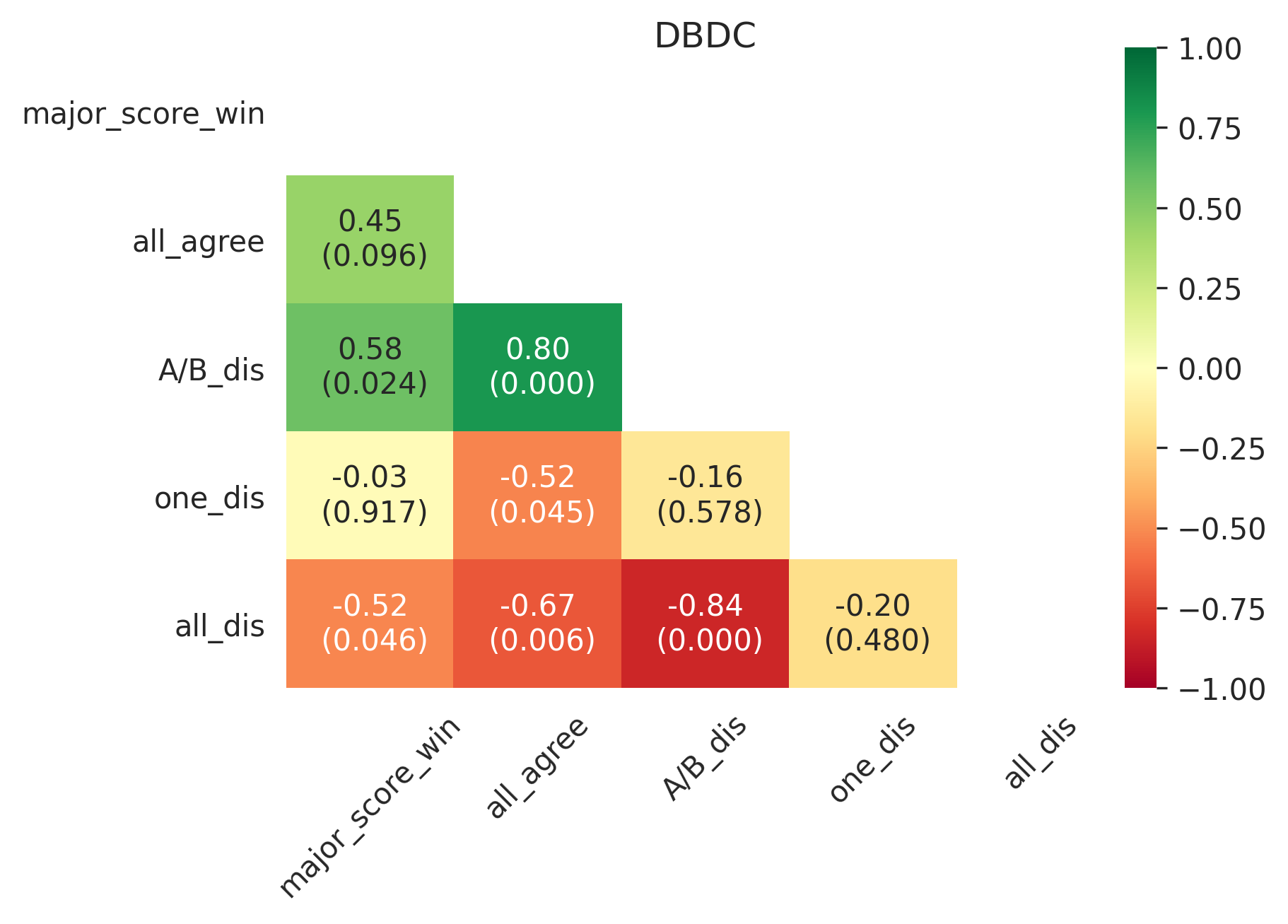}\hfil
\caption{Results on the correlation between $major_{score(win)}$ and agreement \& disagreement features in DBDC.}
\label{fig:dbdc_corr_spear}
\end{figure}

\begin{figure}[H]
\centering
    \includegraphics[width=1\linewidth]{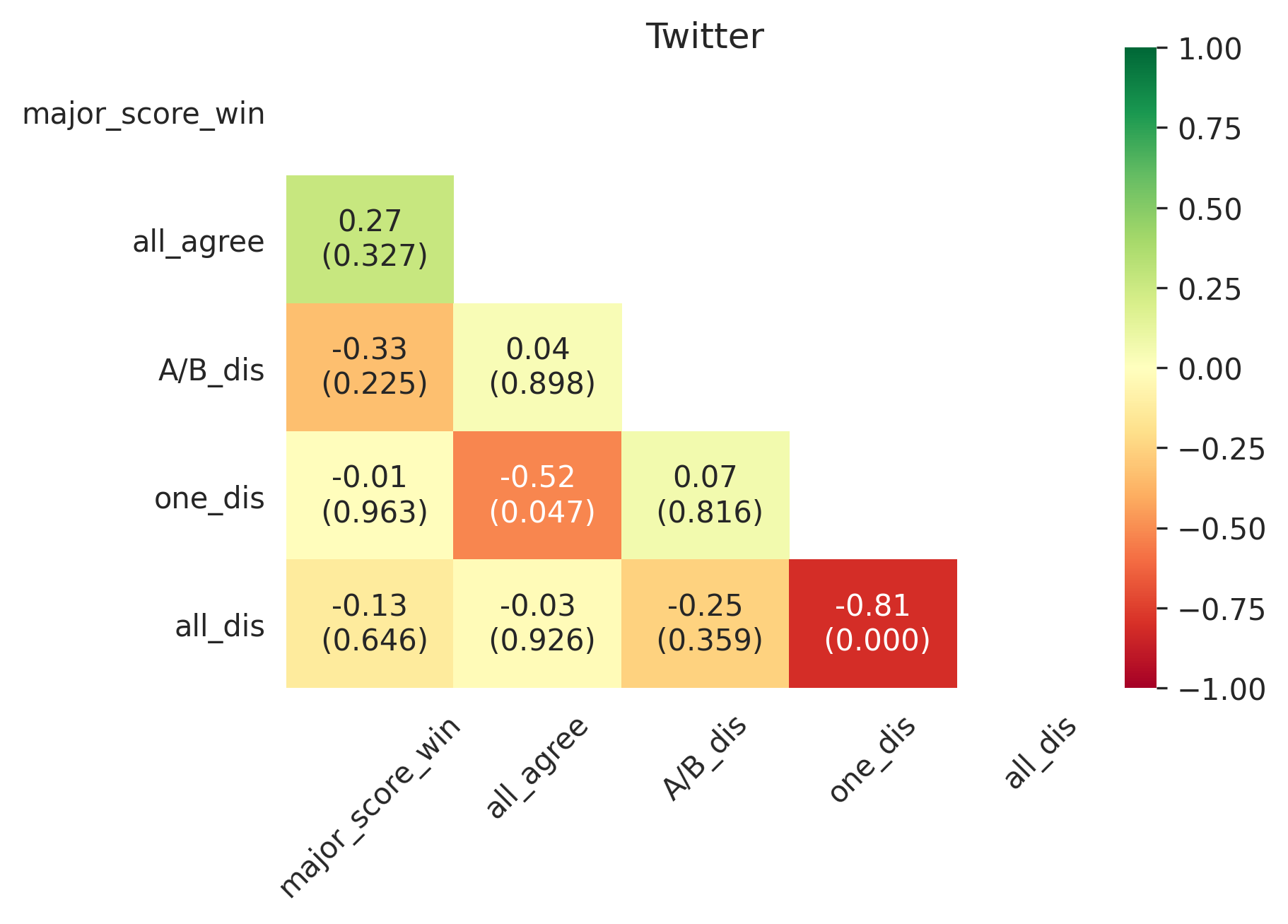}\hfil
\caption{Results on the correlation between $major_{score(win)}$ and agreement \& disagreement features in Twitter.}
\label{fig:twitter_corr_spear}
\end{figure}

\begin{figure}[H]
\centering
    \includegraphics[width=1\linewidth]{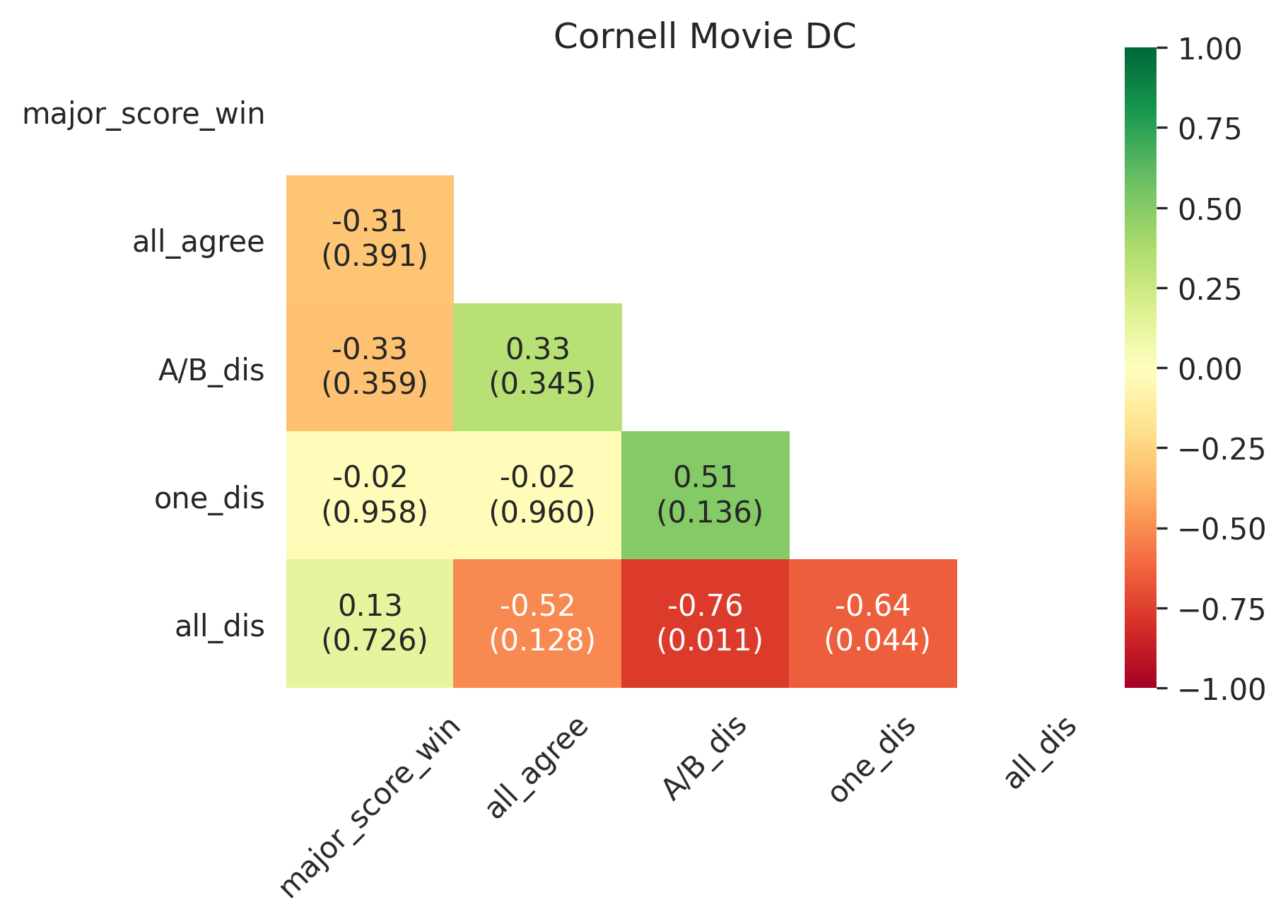}\hfil
\caption{Results on the correlation between $major_{score(win)}$ and agreement \& disagreement features in Cornell Movie DC.}
\label{fig:cornell_corr_spear}
\end{figure}

\begin{figure}[H]
\centering
    \includegraphics[width=1\linewidth]{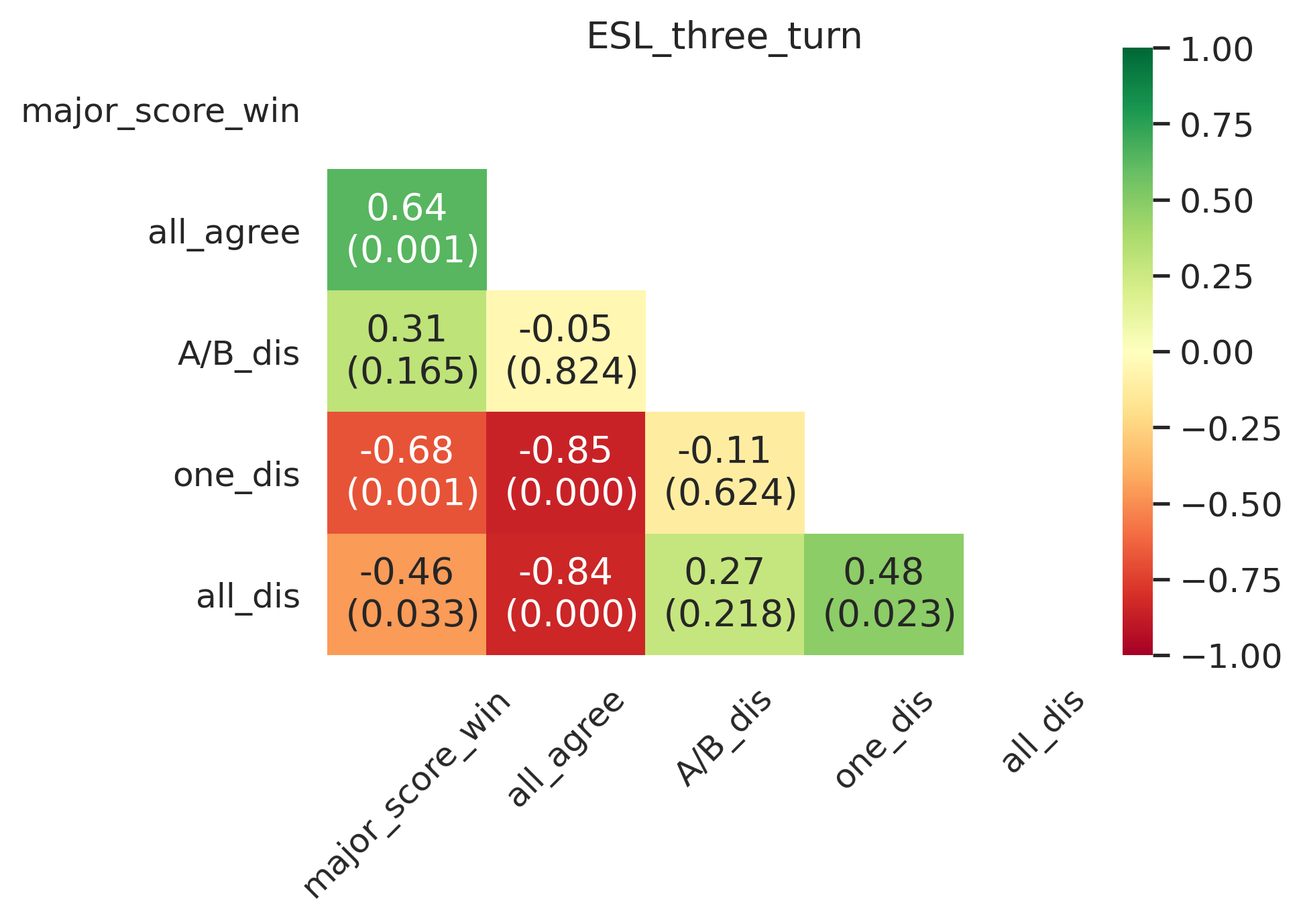}\hfil
\caption{Results on the correlation between $major_{score(win)}$ and agreement \& disagreement features in ESL three turns.}
\label{fig:multi_corr_spear}
\end{figure}

\section{Persona chat with Blender in NCME} \label{appen:persona}

\begin{table}[h]
    \centering
    \scalebox{0.8}{%
    \begin{tabular}{l} \hline
         \textbf{Persona} \\ \hline
            I love chocolate cake with extra frosting.\\
            I have two siberian huskies.  \\ \hline
         \textbf{Example} \\ \hline
            \textbf{Prompt 1:} \textbf{who are you crazy about?}\\
            \textbf{A:} I'm not crazy about anyone in particular. \\ I just want to meet new people \\ and make new friends. (\it{persona-free}) \\
            \textbf{B:} I'm crazy about my dog, Husky. \\ He's my best friend. How about you?\\(\it{with persona})  \\ \hline
            \textbf{Prompt 2:} \textbf{what is the color of a leaf ?}\\
            \textbf{A:} It depends on the type of leaf. Some are white, \\some are yellow, and some are brown. (\it{persona-free})  \\
            \textbf{B:} I'm not sure, but I do know that \\a husky is a sled-type of dog.\\(\it{with persona})  \\ \hline
    \end{tabular}}
    \caption{\label{tab:persona} Persona chat examples with Blender in NCME.}
\end{table}

\onecolumn
\section{Weak Agreement with A/B Results} \label{appen:weak}
We shows workers voting results in Chateval A/B paired test with significant test and weak agreement. $A_{votes}$, $B_{votes}$ and $Tie$ indicates percentage of voting results. We shows Bootstraping test~\cite{dror2018hitchhiker, berg2012empirical} for the significant test with all of the pairwise matchups.

\begin{table*}[h]  
    \centering
    \scalebox{0.6}{
    \begin{tabular}{|c|c|c|c|c|c|c|c|c|c|}
        \hline
         A & B & $A_{votes}$ & $B_{votes}$ & $Tie$ & $all_{agree}$ & $A/B_{dis}$ & $one_{dis}$ & $all_{dis}$ & p-value\\ \hline
          \multirow{9}{*}{NCME human 1} & Blender(2.7B) & \gradient{30}\% & \gradient{40}\% & \gradient{30}\% & 29 & 51 & 130 & 41 & 0.0\\ \cline{2-10}
          & DialoGPT & \gradient{46}\% & \gradient{41}\% & \gradient{13}\% & 3 & 99 & 147 & 22 & 0.07 \\ \cline{2-10}
          & Transformer & \gradient{63}\% & \gradient{21}\% & \gradient{16}\% & 86 & 48 & 99 & 15 & 0.0 \\ \cline{2-10}
          & ParlAI(Controllable) & \gradient{70}\% & \gradient{10}\% & \gradient{20}\% & 103 & 25 & 85	&	11 & 0.0 \\ \cline{2-10}
          & CakeChat & \gradient{61}\% & \gradient{19}\% & \gradient{20}\%  & 87 & 45 & 90 &	23   & 0.0 \\ \cline{2-10}
          & OpenNMT(OS) & \gradient{53}\% & \gradient{21}\% & \gradient{26}\% & 78& 38  & 98&	18  & 0.0 \\ \cline{2-10}
          & OpenNMT(Twitter) & \gradient{74}\% & \gradient{14}\% & \gradient{12}\% & 100 & 34 &82	&17  & 0.0 \\ \cline{2-10} 
          & ConvAI2(seq2seq) & \gradient{70}\% & \gradient{23}\% & \gradient{7}\% & 77 & 82 & 110	&	13 & 0.0 \\ \cline{2-10} \hline
          \multirow{8}{*}{NCME human 2} & Blender(2.7B) & \gradient{34}\% & \gradient{36}\% & \gradient{30}\% & 24 & 52 &128&	48 & 0.27  \\ \cline{2-10}
          & DialoGPT & \gradient{33}\% & \gradient{31}\% & \gradient{36}\%  & 2 & 24 &197	&-  & 0.13\\ \cline{2-10}
          & Transformer & \gradient{35}\% & \gradient{29}\% & \gradient{36}\%  & 20 & 36 &128	&52   & 0.02\\ \cline{2-10}
          & ParlAI(Controllable) & \gradient{39}\% & \gradient{32}\% & \gradient{29}\%  & 16 & 62 & 137&	47 & 0.01 \\ \cline{2-10}
          & CakeChat & \gradient{41}\% & \gradient{33}\% & \gradient{26}\%   & 25 & 56 & 123 &	52 & 0.01 \\ \cline{2-10}
          & OpenNMT(OS) & \gradient{46}\% & \gradient{29}\% & \gradient{25}\% & 28 & 63 & 136	&32   & 0.0 \\ \cline{2-10}
          & OpenNMT(Twitter) & \gradient{45}\% & \gradient{31}\% & \gradient{24}\%   &28 & 61 &141&	31  & 0.0 \\ \cline{2-10} 
          & ConvAI2(seq2seq) & \gradient{44}\% & \gradient{33}\% & \gradient{23}\%   &35 & 56 &124	&41 & 0.0 \\ \cline{2-10} \hline
          \multirow{7}{*}{Blender(2.7B)} & DialoGPT & \gradient{40}\% & \gradient{36}\% & \gradient{24}\% &25 & 66 &141	&34 & 0.16  \\ \cline{2-10} 
          & Transformer & \gradient{39}\% & \gradient{35}\% & \gradient{26}\%  &18 & 57 & 136 &	46 & 0.11 \\ \cline{2-10}
          & ParlAI(Controllable) & \gradient{39}\% & \gradient{36}\% & \gradient{25}\% &30 & 65 &  141	&29 & 0.21 \\ \cline{2-10}
          & CakeChat & \gradient{32}\% & \gradient{28}\% & \gradient{40}\% & 23 & 30 &  122&	55 & 0.06 \\ \cline{2-10}
          & OpenNMT(OS) & \gradient{33}\% & \gradient{30}\% & \gradient{37}\%  & 19 & 34 &142	&39 & 0.19 \\ \cline{2-10}
          & OpenNMT(Twitter) & \gradient{39}\% & \gradient{32}\% & \gradient{29}\% &27 & 52 &134&	39 &  0.04\\ \cline{2-10}
          & ConvAI2(seq2seq) & \gradient{35}\% & \gradient{37}\% & \gradient{28}\%  &23 & 61 & 127&	50 & 0.35 \\ \cline{2-10} \hline
          \multirow{6}{*}{DialoGPT} & Transformer & \gradient{55}\% & \gradient{29}\% & \gradient{16}\% & 53 & 75 & 122 &	25 & 0.0   \\ \cline{2-10}
          & ParlAI(Controllable) & \gradient{58}\% & \gradient{18}\% & \gradient{24}\%  & 73 & 33 & 100 &	27 & 0.0\\ \cline{2-10}
          & CakeChat & \gradient{49}\% & \gradient{33}\% & \gradient{18}\%  & 36 & 75 & 130 &	34 & 0.0 \\ \cline{2-10}
          & OpenNMT(OS) & \gradient{47}\% & \gradient{28}\% & \gradient{25}\%  & 45 & 47 & 112 &	40 & 0.0 \\ \cline{2-10}
          & OpenNMT(Twitter) & \gradient{70}\% & \gradient{14}\% & \gradient{16}\% & 102 & 34 & 87 &	11  & 0.0\\ \cline{2-10}
          & ConvAI2(seq2seq) & \gradient{72}\% & \gradient{19}\% & \gradient{9}\%  & 89 & 68 & 96 &	15 & 0.0 \\ \cline{2-10} \hline
          \multirow{5}{*}{Transformer} & ParlAI(Controllable) & \gradient{40}\% & \gradient{35}\% & \gradient{25}\% & 40 & 57 & 125 & 35& 0.07  \\ \cline{2-10} 
          & CakeChat & \gradient{37}\% & \gradient{36}\% & \gradient{27}\%   & 44 & 45 &123&	33 & 0.39 \\ \cline{2-10}
          & OpenNMT(OS) & \gradient{38}\% & \gradient{41}\% & \gradient{21}\%  & 51 & 58 &122 &	27  & 0.11\\ \cline{2-10}
          & OpenNMT(Twitter) & \gradient{41}\% & \gradient{31}\% & \gradient{28}\%  & 44 & 49 &129&	27  & 0.0 \\ \cline{2-10}
          & ConvAI2(seq2seq) & \gradient{45}\% & \gradient{27}\% & \gradient{28}\%  & 32 & 51 &134	&	34  & 0.0 \\ \cline{2-10} \hline
          \multirow{4}{*}{\shortstack[c]{ParlAI\\(Controllable)}} & CakeChat & \gradient{35}\% & \gradient{40}\% & \gradient{25}\%  & 47 & 53 &123&	30 & 0.08 \\ \cline{2-10} 
          & OpenNMT(OS) & \gradient{31}\% & \gradient{50}\% & \gradient{19}\%   & 54 & 68 &115&	31 & 0.0\\ \cline{2-10}
          & OpenNMT(Twitter) & \gradient{42}\% & \gradient{38}\% & \gradient{20}\% & 43 & 65 &130	&	27 &0.14    \\ \cline{2-10}
          & ConvAI2(seq2seq) & \gradient{39}\% & \gradient{30}\% & \gradient{31}\%  & 43 & 40 &125&	32 &0.0 \\ \cline{2-10} \hline
          \multirow{3}{*}{CakeChat} & OpenNMT(OS) & \gradient{43}\% & \gradient{46}\% & \gradient{11}\% & 47 & 94 & 121	&32 & 0.22 \\ \cline{2-10} 
          & OpenNMT(Twitter) & \gradient{46}\% & \gradient{40}\% & \gradient{14}\%   & 55 & 42 &115&	30 & 0.0 \\ \cline{2-10}
          & ConvAI2(seq2seq) & \gradient{46}\% & \gradient{41}\% & \gradient{13}\%  & 5 & 97 &195&	- & 0.04 \\ \cline{2-10} \hline
          \multirow{2}{*}{\shortstack[c]{OpenNMT\\(OS)}} & OpenNMT(Twitter) & \gradient{43}\% & \gradient{42}\% & \gradient{15}\% & 29 & 96 &139&	32 & 0.36\\ \cline{2-10} 
          & ConvAI2(seq2seq) & \gradient{52}\% & \gradient{35}\% & \gradient{13}\%   & 52 & 87 &125&	23 &0.0\\ \cline{2-10} \hline
          \makecell{OpenNMT\\(Twitter)} & ConvAI2(seq2seq) & \gradient{48}\% & \gradient{37}\% & \gradient{15}\%  & 29 & 92 &140	&31 & 0.0 \\ \cline{2-10} \hline
          Blender(2.7B) & Blender(90M) & \gradient{35}\% & \gradient{30}\% & \gradient{35}\% & 23 & 40 & 131&	46 & 0.10 \\ \cline{2-10} \hline
          Blender(2.7B) & Blender(with persona) & \gradient{44}\% & \gradient{40}\% & \gradient{16}\% & 37 & 80 & 125&	27 & 0.09\\ \cline{2-10} \hline
    \end{tabular}}
        \caption{\label{tab:ncm2} The result of Chateval A/B paired test on NCME. Note that $one_{dis}$ indicates one worker disagree than others, $all_{agree}$ indicates all of workers agrees, $all_{dis}$ indicates all of workers disagrees and $A/B_{dis}$ indicates A/B wins or losses. p-value shows Bootstraping statistical test of A/B paired test (p-value $<$ 0.05 indicates statistically significant).} 
\end{table*}

\begin{table*}[h]
    \centering
    \scalebox{0.55}{
    \begin{tabular}{|c|c|c|c|c|c|c|c|c|c|}
        \hline
         A & B & $A_{votes}$ & $B_{votes}$ & $Tie$ & $all_{agree}$ & $A/B_{dis}$ & $one_{dis}$ & $all_{dis}$ & p-value\\ \hline
          \multirow{5}{*}{Human} & Blender(2.7B) & \gradient{37}\% & \gradient{34}\% & \gradient{29}\% & 15 & 61 &146&	39  &0.15 \\ \cline{2-10} 
          & DialoGPT & \gradient{42}\% & \gradient{46}\% & \gradient{12}\%  & 41 & 95 &139 &	20 &0.18 \\ \cline{2-10}
          & Transformer & \gradient{54}\% & \gradient{36}\% & \gradient{10}\% & 48 & 96 & 127&	25 &0.0\\ \cline{2-10}
          & ParlAI(Controllable) & \gradient{50}\% & \gradient{37}\% & \gradient{13}\%  & 44 & 94 &132&	24 &0.0  \\ \cline{2-10}
          & OpenNMT(OS) & \gradient{51}\% & \gradient{34}\% & \gradient{15}\% & 33 & 93 & 130	&	35  & 0.0\\ \cline{2-10} \hline
          \multirow{4}{*}{Blender(2.7B)} & DialoGPT & \gradient{29}\% & \gradient{30}\% & \gradient{41}\% &  25 & 37 & 137&	38 & 0.41  \\ \cline{2-10} 
          & Transformer & \gradient{31}\% & \gradient{30}\% & \gradient{39}\%   & 27 & 29 &130&	43  & 0.47\\ \cline{2-10}
          & ParlAI(Controllable) & \gradient{31}\% & \gradient{30}\% & \gradient{39}\% & 28 & 34 & 130&	42 & 0.35 \\ \cline{2-10}
          & OpenNMT(OS) & \gradient{31}\% & \gradient{27}\% & \gradient{42}\%  & 21 & 30 &144	&	35 & 0.09  \\ \cline{2-10} \hline
          \multirow{3}{*}{DialoGPT} & Transformer & \gradient{46}\% & \gradient{35}\% & \gradient{19}\%   & 33 & 78 &133&	33 & 0.0 \\ \cline{2-10}
          & ParlAI(Controllable) & \gradient{51}\% & \gradient{35}\% & \gradient{14}\%   & 32 & 93 &134&	34 & 0.36 \\ \cline{2-10}
          & OpenNMT(OS) & \gradient{47}\% & \gradient{31}\% & \gradient{22}\%   & 33 & 71 &140	&27  & 0.09\\ \cline{2-10} \hline
          \multirow{2}{*}{Transformer} & ParlAI(Controllable) & \gradient{40}\% & \gradient{41}\% & \gradient{19}\% & 27 & 78 & 142&	31  & 0.46 \\ \cline{2-10} 
          & OpenNMT(OS) & \gradient{39}\% & \gradient{39}\% & \gradient{22}\%  & 34 & 64 & 130	&36 & 0.40 \\ \cline{2-10} \hline
          \makecell{ParlAI\\(Controllable)} & OpenNMT(OS) & \gradient{41}\% & \gradient{37}\% & \gradient{22}\% & 29 & 62  &131&	40 & 0.14\\ \cline{2-10} \hline
    \end{tabular}}
        \caption{\label{tab:dbdc2} The result of Chateval A/B paired test on DBDC.  }
\end{table*}

\begin{table*}[h]
    \centering
    \scalebox{0.6}{
    \begin{tabular}{|c|c|c|c|c|c|c|c|c|c|}
        \hline
         A & B & $A_{votes}$ & $B_{votes}$ & $Tie$ & $all_{agree}$ & $A/B_{dis}$ & $one_{dis}$ & $all_{dis}$ & p-value \\ \hline
          \multirow{5}{*}{\shortstack[c]{Twitter\\Baseline}} & Blender(2.7B) & \gradient{30}\% & \gradient{33}\% & \gradient{38}\% & 19 & 40 & 138&	43  & 0.16  \\ \cline{2-10} 
          & DialoGPT & \gradient{32}\% & \gradient{41}\% & \gradient{27}\% & 22 & 56 & 138&	40  & 0.0  \\ \cline{2-10}
          & Transformer & \gradient{38}\% & \gradient{37}\% & \gradient{25}\%  & 30 & 61 & 129&	41  & 0.41\\ \cline{2-10}
          & ParlAI(Controllable) & \gradient{35}\% & \gradient{33}\% & \gradient{32}\%  & 23 & 49 &133&	44 & 0.28 \\ \cline{2-10}
          & ConvAI2(seq2seq) & \gradient{36}\% & \gradient{36}\% & \gradient{28}\%  & 26 & 55 & 124	&50 & 0.46 \\ \cline{2-10} \hline
          \multirow{4}{*}{Blender(2.7B)} & DialoGPT & \gradient{32}\% & \gradient{32}\% & \gradient{36}\%  & 24 & 41 & 139	&	37 & 0.47 \\ \cline{2-10}
          & Transformer & \gradient{34}\% & \gradient{35}\% & \gradient{31}\% & 23 & 49 & 133&	44 & 0.40 \\ \cline{2-10}
          & ParlAI(Controllable) & \gradient{28}\% & \gradient{33}\% & \gradient{39}\%  & 28 & 33 & 131	&	41 & 0.06\\ \cline{2-10}
          & ConvAI2(seq2seq) & \gradient{30}\% & \gradient{29}\% & \gradient{41}\% & 20 & 33 & 133&	47 & 0.45 \\ \cline{2-10} \hline
          \multirow{3}{*}{DialoGPT} & Transformer & \gradient{36}\% & \gradient{32}\% & \gradient{32}\%  & 26 & 50 & 130&	44 & 0.07  \\ \cline{2-10}
          & ParlAI(Controllable) & \gradient{36}\% & \gradient{36}\% & \gradient{28}\%  & 18 & 56 & 147	&35 & 0.48 \\ \cline{2-10}
          & ConvAI2(seq2seq) & \gradient{38}\% & \gradient{28}\% & \gradient{34}\% & 27 & 47 &  137	&	36  & 0.0 \\ \cline{2-10} \hline
          \multirow{2}{*}{Transformer} & ParlAI(Controllable) & \gradient{33}\% & \gradient{35}\% & \gradient{32}\%  & 26 & 42 & 126&	48 & 0.27 \\ \cline{2-10} 
          & ConvAI2(seq2seq) & \gradient{39}\% & \gradient{25}\% & \gradient{36}\%  & 24 & 29 & 128&	48 & 0.0\\ \cline{2-10}\hline
          \makecell{ParlAI\\(Controllable)} & ConvAI2(seq2seq) & \gradient{41}\% & \gradient{29}\% & \gradient{30}\% & 27 & 46 & 136&	37  & 0.0 \\ \cline{2-10} \hline
    \end{tabular}}
        \caption{\label{tab:twitter2} The result of Chateval A/B paired test on Twitter. }
\end{table*}

\begin{table*}[h]
    \centering
    \scalebox{0.6}{
    \begin{tabular}{|c|c|c|c|c|c|c|c|c|c|}
        \hline
         A & B & $A_{votes}$ & $B_{votes}$ & $Tie$ & $all_{agree}$ & $A/B_{dis}$ & $one_{dis}$ & $all_{dis}$ & p-value\\ \hline
          \multirow{4}{*}{\shortstack[c]{CornellMovie \\ DC Baseline}} & Blender(2.7B) & \gradientnew{40}\% & \gradientnew{41}\% & \gradientnew{19}\% & 147 & 396 &675&	178 & 0.32 \\ \cline{2-10} 
          & DialoGPT & \gradientnew{39}\% & \gradientnew{39}\% & \gradientnew{22}\%  & 142 & 345 &646&	194 & 0.38 \\ \cline{2-10}
          & DC-NeuralConv & \gradientnew{41}\% & \gradientnew{40}\% & \gradientnew{19}\%  & 119 & 411 &687&	212 & 0.06  \\ \cline{2-10}
          & ParlAI(Controllable) & \gradientnew{42}\% & \gradientnew{40}\% & \gradientnew{18}\%  & 133 & 416 &683&	184 &0.03 \\ \cline{2-10} \hline
          \multirow{3}{*}{Blender(2.7B)} & DialoGPT & \gradientnew{36}\% & \gradientnew{38}\% & \gradientnew{26}\%  & 124 & 290 &652	&	224 & 0.18 \\ \cline{2-10}
          & DC-NeuralConv & \gradientnew{43}\% & \gradientnew{33}\% & \gradientnew{24}\% & 144 & 300 & 647	&209 & 0.0 \\ \cline{2-10}
          & ParlAI(Controllable) & \gradientnew{39}\% & \gradientnew{37}\% & \gradientnew{24}\%  & 132 & 315 &655&	213 & 0.08 \\ \cline{2-10}\hline
          \multirow{2}{*}{DialoGPT} & DC-NeuralConv & \gradientnew{38}\% & \gradientnew{36}\% & \gradientnew{26}\% & 109 & 292 &673	&	218  & 0.01 \\ \cline{2-10}
          & ParlAI(Controllable) & \gradientnew{40}\% & \gradientnew{38}\% & \gradientnew{22}\%   & 210 & 343 &687&	193 & 0.12\\ \cline{2-9}\hline
          \makecell{DC-\\NeuralConv} & ParlAI(Controllable) & \gradientnew{39}\% & \gradientnew{35}\% & \gradientnew{26}\% & 122 & 292 &670&	208 & 0.01  \\ \cline{2-10} \hline
    \end{tabular}}
        \caption{\label{tab:cornell2} The result of Chateval A/B paired test on Cornell Movie DC. }
\end{table*}

\begin{table*}[h]
    \centering
    \scalebox{0.6}{
    \begin{tabular}{|c|c|c|c|c|c|c|c|c|c|}
        \hline
         A & B & $A_{votes}$ & $B_{votes}$ & $Tie$ & $all_{agree}$ & $A/B_{dis}$ & $one_{dis}$ & $all_{dis}$ & p-value\\ \hline
          \multirow{6}{*}{Human} & Blender(2.7B) & \gradient{38}\% & \gradient{21}\% & \gradient{41}\% & 36 & 21 & 122&	42 &  0.0 \\ \cline{2-10} 
          & DialoGPT & \gradient{58}\% & \gradient{18}\% & \gradient{24}\% & 42 & 50 & 127&	31  & 0.0  \\ \cline{2-10}
          & CakeChat & \gradient{60}\% & \gradient{18}\% & \gradient{22}\% & 50 & 60 & 125&	25  & 0.0  \\ \cline{2-10}
          & ParlAI(Controllable) & \gradient{55}\% & \gradient{17}\% & \gradient{28}\%  & 47 & 39 & 125&	25 & 0.0 \\ \cline{2-10}
          & ConvAI2(KV-MemNN) & \gradient{66}\% & \gradient{13}\% & \gradient{21}\%  & 63 & 38 &109&	28 & 0.0 \\ \cline{2-10}
          & ConvAI2(seq2seq) & \gradient{66}\% & \gradient{11}\% & \gradient{23}\%  & 62 & 38 & 121	& 17 & 0.0 \\ \cline{2-10} \hline
          \multirow{6}{*}{Blender(2.7B)} & BlenderPersona & \gradient{50}\% & \gradient{17}\% & \gradient{33}\%  & 61 & 24 & 115	&	20 & 0.0 \\ \cline{2-10}
          & DialoGPT & \gradient{79}\% & \gradient{10}\% & \gradient{11}\% & 120 & 31 & 70&	10  & 0.0 \\ \cline{2-10}
          & CakeChat & \gradient{60}\% & \gradient{11}\% & \gradient{29}\%  & 85 & 26 & 99	&	16 & 0.0\\ \cline{2-10}
          & ParlAI(Controllable) & \gradient{56}\% & \gradient{16}\% & \gradient{28}\% & 61 & 20 & 113&	26 & 0.0 \\ \cline{2-10} 
          & ConvAI2(KV-MemNN) & \gradient{64}\% & \gradient{10}\% & \gradient{26}\% & 71 & 27 & 113 &	16 & 0.0 \\ \cline{2-10} 
          & ConvAI2(seq2seq) & \gradient{60}\% & \gradient{5}\% & \gradient{35}\% & 74 & 11 & 121 &	5 & 0.0 \\ \cline{2-10} \hline
          \multirow{4}{*}{DialoGPT} & ConvAI2(KV-MemNN) & \gradient{25}\% & \gradient{16}\% & \gradient{59}\%  & 53 & 10 & 132 &	15  & 0.0 \\ \cline{2-10}
          & ConvAI2(seq2seq) & \gradient{33}\% & \gradient{24}\% & \gradient{43}\%  & 47 & 20 & 133	& 20 & 0.0 \\ \cline{2-10}
          & ParlAI(Controllable) & \gradient{18}\% & \gradient{25}\% & \gradient{57}\% & 67 & 5 &  115	&	36 & 0.01 \\ \cline{2-10}
          & CakeChat & \gradient{29}\% & \gradient{22}\% & \gradient{49}\% & 49 & 13 &  128	&	23  & 0.01\\ \cline{2-10} \hline
          \multirow{3}{*}{CakeChat} & ParlAI(Controllable) & \gradient{28}\% & \gradient{46}\% & \gradient{26}\%  & 53 & 48 & 123&	24 & 0.0 \\ \cline{2-10} 
          & ConvAI2(KV-MemNN) & \gradient{35}\% & \gradient{25}\% & \gradient{40}\%  & 34 & 26 & 137	& 29 & 0.0\\ \cline{2-10}
          & ConvAI2(seq2seq) & \gradient{35}\% & \gradient{30}\% & \gradient{35}\%  & 37 & 33 & 130 &	33 & 0.07 \\ \cline{2-10}\hline
          \makecell{ParlAI\\(Controllable)} & ConvAI2(KV-MemNN) & \gradient{48}\% & \gradient{17}\% & \gradient{35}\% & 27 & 46 & 136&	37  & 0.0 \\ \cline{2-10} 
          & ConvAI2(seq2seq) & \gradient{38}\% & \gradient{20}\% & \gradient{42}\%  & 33 & 20 & 134	& 32 & 0.0\\ \cline{2-10} \hline
          \makecell{ConvAI2\\(KV-MemNN)} & ConvAI2(seq2seq) & \gradient{23}\% & \gradient{32}\% & \gradient{45}\% & 36 & 25 & 132 &	32  & 0.0 \\ \cline{2-10} \hline
    \end{tabular}}
        \caption{\label{tab:esl} The result of Chateval A/B paired test on ESL 3-turns.}
\end{table*}

\end{document}